%% file: main.tex
\documentclass{article}

\usepackage{subfiles}
\usepackage[preprint]{neurips_2025}

\usepackage{xcolor}
\usepackage[utf8]{inputenc} 
\usepackage[T1]{fontenc}    
\usepackage{hyperref}       
\usepackage{url}            
\usepackage{booktabs}       
\usepackage{amsfonts}       
\usepackage{nicefrac}       
\usepackage{microtype}      
\usepackage{xcolor}         
\usepackage{amsmath}
\usepackage{float}
\usepackage{hyperref}
\usepackage{graphicx}
\graphicspath{{figures/}}  
\usepackage{caption}
\usepackage{subcaption}
\usepackage{amsthm}
\usepackage{natbib}
\usepackage{pdfpages}  
\usepackage{pax}

\newtheorem{proposition}{Proposition}
\newtheorem{theorem}{Theorem}
\newtheorem{corr}{Corollary}      
\newtheorem{observation}{Observation}[section]

\newcommand{\R}{\mathbb{R}}
\renewcommand{\d}{\text{d}}

\newcommand{\Err}{\text{Err}}
\newcommand{\E}{\mathbb{E}}

\newcommand{\todo}[1]{}

\newcommand{\diag}{\text{diag}}

\DeclareMathOperator{\Id}{Id}
\renewcommand{\d}{\text{d}}
\newcommand{\K}{\mathcal{K}}
\newcommand{\T}{\mathbb{T}}
\renewcommand{\P}{\mathcal{P}}
\newcommand{\smallmultiline}[1]{\parbox{\linewidth}{ \begin{center}{\normalsize \textit{#1}}\end{center} } }

\newcommand{\EcommentDONE}[1]{}

\title{KPFlow: An Operator Perspective on Dynamic Collapse Under Gradient Descent Training of Recurrent Networks}
\author{James Hazelden$^{1,4,5*}$, Laura Driscoll$^{2,4,5\dagger}$, Eli Shlizerman$^{1,3,4*}$, Eric Shea-Brown$^{1,4,5*}$\\
\smallmultiline{
    $^1$Department of Applied Mathematics, $^2$Department of Neurobiology \& Biophysics, $^3$Department of Electrical \& Computer Engineering, $^4$Computational Neuroscience Center\\
    University of Washington, Seattle, WA 98195, \\ 
    \vspace{.3cm}
    $^5$Allen Institute, Seattle, WA 98109, \\ 
    *\texttt{\{jhazelde,shlizee,etsb\}@uw.edu}, $\, \dagger$\texttt{laura.driscoll@alleninstitute.org}
}
}
\date{}

\begin{document}

\maketitle


\vspace{-.55cm}

\begin{abstract}
Gradient Descent (GD) and its variants are the primary tool for enabling efficient training of recurrent dynamical systems such as Recurrent Neural Networks (RNNs), Neural ODEs and Gated Recurrent units (GRUs). The dynamics that are formed in these models exhibit features such as neural collapse and emergence of latent representations that may support the remarkable generalization properties of networks. In neuroscience, qualitative features of these representations are used to compare learning in biological and artificial systems. Despite recent progress, there remains a need for theoretical tools to rigorously understand the mechanisms shaping learned representations, especially in finite, non-linear models. Here, we show that the gradient flow, which describes how the model's dynamics evolve over GD, can be decomposed into a product that involves two operators: a Parameter Operator, $\K$, and a Linearized Flow Propagator, $\P$. $\K$ mirrors the Neural Tangent Kernel in feed-forward neural networks, while $\P$ appears in Lyapunov stability and optimal control theory. We demonstrate two applications of our decomposition. First, we show how their interplay gives rise to low-dimensional latent dynamics under GD, and, specifically, how the collapse is a result of the network structure, over and above the nature of the underlying task. Second, for multi-task training, we show that the operators can be used to measure how objectives relevant to individual sub-tasks align. We experimentally and theoretically validate these findings, providing an efficient Pytorch package, \emph{KPFlow}, implementing robust analysis tools for general recurrent architectures. Taken together, our work moves towards building a next stage of understanding of GD learning in non-linear recurrent models.
\end{abstract}

\section{Introduction}

Gradient descent (GD) and its accelerated or stochastic variants have been demonstrated in deep learning and, more classically, optimal control, to be an effective means of training a parameterized dynamical systems to align with a particular task, consisting of inputs and target behavior for the model \citep{chen2018neural, lecun2015deep}. When such a model is trained with GD, the parameters are iteratively modified in such a way that the averaged loss over all input conditions (referred to as trials throughout this work) is minimized. In prior work, it has been observed that in single- and multi-task contexts, parameterized dynamical systems such as recurrent neural networks (RNNs) trained with GD exhibit a \emph{dynamic collapse} to interpretable low-dimensional solutions consisting of fixed-point attractors and other dynamical motifs \citep{sussillo2013opening, farrell2023from, driscoll2024flexible, turner2023simplicity, NEURIPS2020sch}. While these works have provided a comprehensive exploration of various aspects of this collapse, there is a gap in formulation and understanding of why GD exhibits a bias towards dynamical collapse, especially in the realistic case of non-linear dynamical models with finitely many neurons. A comprehensive discussion of prior related work is in Appendix~\ref{sec:ap1}. 

In this work, we propose a decomposition, \emph{KPFlow}, of each stage of GD learning in recurrent models as the application of individual compact operators that act on the space of per-trial trajectories, which are 3-tensors over all forward-pass times, $t$, task input trials, $x$, and hidden units. In particular, our decomposition matches the true gradient flow of the hidden state of the model as it is iteratively adjusted by GD. In addition, we theoretically analyze the operators and their spectral decomposition and show how they can be practically used to develop a deeper view of each stage of GD. Finally, we provide a Pytorch package \citep{paszke2019pytorchimperativestylehighperformance} efficiently implementing the operators for general recurrent models. Ultimately, we use the tools established to provide a rigorous exploration of the mechanisms leading to dynamic collapse and attractor re-use in non-linear recurrent models.

Our \textit{primary contributions} include:

\begin{enumerate}
    \item Proposition \ref{thm:gdflow} derives the KPFlow decomposition of the gradient flow in terms of two operators, $\P$, the \emph{linearized flow propagator} and $\K$, the \emph{parameter operator}. We describe the implementation details and provide examples. 

    \item In Section \ref{sec:dyncollapse}, we investigate dynamic collapse of RNNs and GRUs with varied initial recurrent weight scales using the KPFlow decomposition. We find that the $\K$ operator bottle-necks the dimension of dynamical changes in all cases, leading to dimension collapse and faster training when $\K$ is higher rank, corresponding to larger initial weight scale. 

    \item Section \ref{sec:interferencemulti} shows that $\P$ is block-diagonal over task trial. When the model is simultaneously trained on multiple tasks, the KPFlow further decomposes into operators quantifying alignment and interference between each sub-task. The metrics developed illustrate in the first few GD iterations which groups of tasks will form aligned activity subspaces.

    \item Section \ref{sec:theory} establishes the relationship between the effective rank of $\K$ and the effective dimension of the recurrent activity for a general ``weight based'' dynamical system. We decompose $\P$ into operators encountered in Lyapunov stability theory, observing that the SVD is high rank.

    \item Finally, our Appendix provides a comprehensive description of the methods, theory and computational implementation. As a diverse illustration, we derive the KPFlow for a network of biophysical \emph{Hodgkin Huxley} neurons \citep{hodgkin1952quantitative}, motivating why truncated gradients may be more effective in this context.
\end{enumerate}



\section{Preliminaries}

\label{sec:prelim}


\subsection{Notation}

A wide range of tasks in deep learning can be defined through the lens of training a parameterized dynamical system to minimize an objective over a distribution of (potentially infinite) trials. These models have a time variable, $t$, that may be discrete, such as for RNNs or GRUs \citep{rumelhart_hinton_williams_1986, bahdanau2016neuralmachinetranslationjointly} or continuous, such as for neural ODEs \citep{chen2018neural}. Our methods are presented in the continuous context for generality, but a discrete version is provided in Appendix~\ref{sec:ap221}. 

We let $X$ denote the task trial distribution. In the supervised context, a trial consists of an input $x(t)$ over all $t$ in a trial-dependent time window $[0, t_{end}]$, and a task target, $y^*(t)$, corresponding to the input $x(t)$. We let $\T$ denote the space of all hidden space per-trial trajectories. When discretized with $H$ hidden neurons, $T$ timesteps, and a batch $B$ of input trials, $\T$ consists of all 3-tensors of shape $[B, T, H]$, but theoretically we don't perform this discretization (see Appendix~\ref{sec:ap22}). Throughout this work, we let $q \in \T$ denote \textit{any} generic per-trial trajectory from $\T$ and use the notation $q(t|x)$ to denote the value of $q$ at time $t$ on trial $x \sim X$. The model's hidden state is a \textit{specific} sample from $\T$, which we denote by $z \in \T$. We assume $z$ is a dynamical system with parameters $\theta$, trained by GD,
\begin{align}
    \label{eqn:forwarddyn}
    \frac{\d}{\d t} z(t|x) &= f(z(t|x), x(t), t | \theta), \, \, z(t|0) = z_0.
\end{align}
The value of $f$ for all $t$ and $x$ is itself an element of $\T$, describing the tangents to $z$ on all trials. 

Finally, an \emph{operator} is a linear mapping from $\T \rightarrow \T$. The KPFlow is based on such operators. In Appendix~\ref{sec:consensus} we describe approaches for simplifying these operators, carefully averaging over all trials, times or space, while respecting the subspace structure, e.g., to measure a trial-averaged SVD. 

\subsection{The Hidden State Gradient Flow}

\begin{figure}[t]
  \centering
  \includegraphics[width=0.85\linewidth]{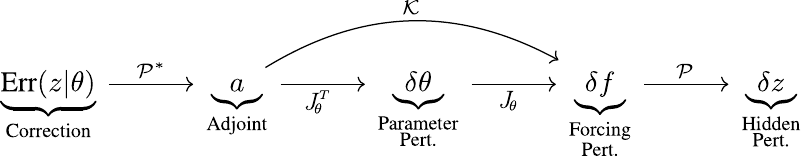}
  \caption{\textbf{Schematic of KPFlow decomposition illustrating the process of transforming error signals into hidden state perturbations.} The error signal $\Err(z|\theta)$ depends on the model parameters $\theta$, and the model hidden state, $z$. Each stage of backpropagation and its eventual impact on the hidden state dynamics, $\delta z$, is described by an operator acting on a space of 3-tensors, with rich properties encoding the structure of learning, as detailed in the main text.}
  \label{fig:gdflow}
\end{figure}

In this section, we address how GD perturbs the model in Equation \ref{eqn:forwarddyn} to align with a given task. Mechanistically, this consists of sequence of operators, summarized in Proposition \ref{thm:gdflow}. 


\paragraph{A single GD iteration} GD backpropagation consists of two steps: (1) adjoint backpropagation and (2) parameter backpropagation \citep{chen2018neural}. First, the adjoint, $a \in \T$, is defined as the gradient of the overall loss $L$ with respect to the hidden state, $a(t|x) := \nabla_{z(t|x)} L$. Practically, this is backpropagated through the dynamical system 
\begin{align}
    \label{eqn:step1}
    \frac{\d}{\d t} a(t|x) &= -J_z(t|x)^T a(t|x) - \Err(t | x), \,\ \, a(t|t_{end}) = 0,
\end{align}
where $\Err \in \T$ is an error signal and $J_z$ is the hidden state Jacobian (see Appendix~\ref{sec:ap21}). Next, a perturbation to the parameters, $\delta  \theta$, is backpropagated by 
\begin{align}
    \label{eqn:step2}
    \delta \theta &= \underset{x_0 \sim X}{\E} \int_0^{t_{end}} J_\theta(t_0 | x_0)^T a(t_0 | x_0) \, \d t_0. 
\end{align}

\paragraph{Perturbations to hidden state} The hidden state, $z \in \T$, is perturbed at each iteration of GD. A generic perturbation $\delta f \in \T$ to the tangential dynamics $f$ in Equation \ref{eqn:forwarddyn} induces a perturbation to $z$ given by the \textit{linearized flow}, reminiscent of the adjoint dynamics, 
\begin{align}
    \label{eqn:step3}
    \delta z(t|x) &= J_z(t|x) \delta z(t|x) + \delta f(t|x), \, \, \delta z(0|x) = 0.
\end{align}

\paragraph{Informally defining the KP operators} The process of producing $\delta f$ from an error signal $\Err \in \T$ over all trials, times and hidden units, is described by two operators. First, we define $\P : \T \rightarrow \T$ by the mapping from $\delta f \in \T$ to $\delta z \in \T$, solving Equation \ref{eqn:step3} forward in time on all trials. Then $\P^*$, the Hermitian adjoint operator, corresponds to the map from $\Err$ to the adjoint in Equation \ref{eqn:step1}. Secondly, we define $\K : \T \rightarrow \T$ as the operator mapping the adjoint to $\delta f$, intermediately solving Equation \ref{eqn:step2}. $\P$ and $\K$ are formally defined in Appendix~\ref{sec:ap22}. Conceptually, $\P$ is classically encountered in Lyapunov stability and controllability analysis in optimal control \citep{nijmeijer1990nonlinear}, while $\K$ generalizes the classical Neural Tangent Kernel (NTK), concentrated to the tangents $f(z) \in \T$ \citep{jacot2018neural}. The following Proposition decomposes the GD flow using these particular choices of operators, effectively unifying NTK theory and classical non-linear control.

\begin{proposition}{KP Gradient Flow Decomposition\\}
    \label{thm:gdflow} 
    With the operators defined as above, the gradient flow translating an error signal, $\Err(t|x) \in \T$, to perturbations to the hidden state, $\delta z(t|x) \in \T$, is given by a composition of three linear operators:
    \begin{align}
        \delta z &= - [\P \circ \K \circ \P^*] (\Err).
    \end{align}
    
\end{proposition}

A proof is in Appendix~\ref{sec:appendixproofflow}. Figure \ref{fig:gdflow} illustrates the process of producing $\delta z$ from $\Err$. The operators $\P$ and $\K$ can dramatically simplify depending on the choice of model or dynamics it exhibits. The rest of our work focuses on analysis and applications of this decomposition. 

\paragraph{Similarities and Differences with NTK} We briefly note that Proposition \ref{thm:gdflow} bears a strong resemblance to the neural tangent kernel (NTK) formulation first presented in \citep{jacot2018neural} and extended by others specifically to single- and multi-layer RNNs \citep{alemohammad2020recurrent}. Other work has formed the NTK for specific neural ODEs by taking the infinite width limit of the NTK of ResNets \citep{gao2025exploring, he2015deepresiduallearningimage}. This said, there are multiple factors that distinguish our decomposition from these prior works. In particular, we decompose into the two specific operators $\P$ and $\K$. We show in the rest of this work that this organization is logical since $\K$ is the only source of mixing between trials and is low rank, as contrasted with $\P$. Furthermore, both operators can be studied and simplified in isolation. Our formulation also differs in its generality, being applicable to any dynamical system model. Our focus also differs from typical NTK applications. While NTK is commonly used to show that gradient descent on neural networks resembles kernel gradient descent in the infinite-width limit, we instead analyze finite networks by leveraging the empirical operators and their SVD. Our approach examines how task specific dynamics emerge under GD and provides a computational package built specifically around this decomposition. See the Appendix for more details on this distinction, as well as an extended Related Work. The following Corollary illustrates how our decomposition can be used to derive an operator-based formulation of model output gradient flow in the generic context of Equation \ref{eqn:forwarddyn}, with a proof in the Appendix including an exploration of the case when $W_{out}$ is trained. In this work, the output weights are always trained with GD.

\begin{corr}{Output GD Flow of General Models\\} 
    Using a fixed linear readout $y(t|x) = W_{out} z(t|x) + b_{out}$ and a squared error loss, $y$ evolves according to an operator $\Theta$ given by
    \begin{align}
        \delta y &= -\Theta (y - y^*) = -[W_{out} \cdot \P \circ \K \circ \P^* \cdot W_{out}^T] (y - y^*).
    \end{align}
\end{corr}
%
%

\paragraph{Practical Implementation} Appendix ~\ref{sec:corrproof} provides a proof of the above. In Appendix~\ref{sec:oppractical}, we describe the multiple ways the introduced KPFlow package implements the operators $\K$ and $\P$ efficiently for generic architectures. Our code can compute the full and reduced SVD of any operators mapping $\T \rightarrow \T$ and measure alignment with arbitrary trajectories, $q \in \T$. For example, since $\K$ is positive definite (see Theorem \ref{thm:operator} below) we can decompose it via  
\begin{align}
    \K q &= \lambda_1 v_1 \langle q, v_1 \rangle + \lambda_2 v_2 \langle q, v_2 \rangle + \cdots,
\end{align}
where $\lambda_1 > \lambda_2 > \cdots \geq0 $ and the inner product on $\T$ averages over task trials and integrates over time and hidden units. The modes $v_i$ encode the structure of $\K$ and encode a huge amount of information. For example, $v_1$ specifies how to maximally stimulate $\K$ at every instant in time, hidden unit and on every input trial $x$. We also introduce a reduced consensus SVD, measuring, for example, the dominant directions in hidden space where $\K$ acts when averaged over trials and times. This formulation is described in Appendix~\ref{sec:consensus} and this consensus SVD is used to measure and bound the effective rank and spectrum of the operators throughout this work. 

\subparagraph{Computing Actions of $\P$ and $\K$} Internally, our code uses matrix-free approaches to compute the SVD and action of the operators, since explicitly discretizing $\P$ and $\K$ is memory intensive. The action of $\P$ can be computed by propagating the linearized flow, Equation \ref{eqn:step3}, forward in time, similarly to computing Lyapunov exponents \citep{vogt2022lyapunov}. Alternatively, $\P \delta f$ can be computed by replacing $f$ with $f + \delta f$ in Equation \ref{eqn:forwarddyn}; both methods agree for small $\delta f$. On the other hand, $\K$ can be efficiently computed without ever explicitly forming the Jacobians $J_\theta$ using the built-in Pytorch functions \textit{Jacobian-vector product} and \textit{vector-jacobian product} \citep{paszke2019pytorchimperativestylehighperformance} (see Appendix~\ref{sec:oppractical}). 

\paragraph{RNN Example} One-layer RNNs are a well-studied model in machine learning and neuroscience \citep{sompolinsky1988chaos, rajan2006eigenvalue, sussillo2009generating, sussillo2013opening}, and serve as a relatively simple example of a non-linear sequential model that illustrates the main points of Proposition \ref{thm:operator}. The RNN has dynamics
\begin{align}
    \label{eqn:rnn}
    f(z(t|x), x(t) \, | \, \{W, W_{in} \}) = -z(t|x) + W \sigma(z(t|x)) + W_{in} x(t).
\end{align}
In this case, the Jacobian is $W \text{diag}(\sigma'(z(t)))$ and $\P$'s action consists of many applications of this Jacobian to propagate a trajectory along the linearized flow. The adjoint operator $\P^*$ instead multiplies by $\text{diag}(\sigma'(z(t))) \cdot W^T$. The $\K$ operator for the RNN is defined as the kernel operator
\begin{align}
    (\K q)(t|x) &= \underset{x_0 \sim X}{\E} \Big[\int_0^{t_{end}} k(x, t, x_0, t_0) \, q(t_0|x_0) \d t_0 \Big],  \\
    \text{ where } \, k(x, t, x_0, t_0) &= x(t)^Tx_0(t_0) + \sigma(z(t|x))^T\sigma(z(t_0|x_0)).
\end{align}
Effectively this averages over trials and integrates over time, weighting by input and recurrent activity alignment. The input contribution comes from training $W_{in}$ and the recurrent term comes from training $W$. Explicitly, $\K$ above is a integral operator of Hilbert-Schmidt type with effective rank bounded above by the sum of the effective dimensions of $x(t)$ and $\sigma(z(t|x))$ over all $t$ and $x$, foreshadowing the more general Theorem \ref{thm:operator} below. In the case of a GRU or LSTM network, the tangential dynamics $f$ are a feed-forward neural network and $\K$ is given by the classical NTK for this network. On the other hand, the operator $\P$ corresponds to the linearized recurrent dynamics. 

\begin{figure}[t]
    \centering
    \includegraphics[width=.95\linewidth]{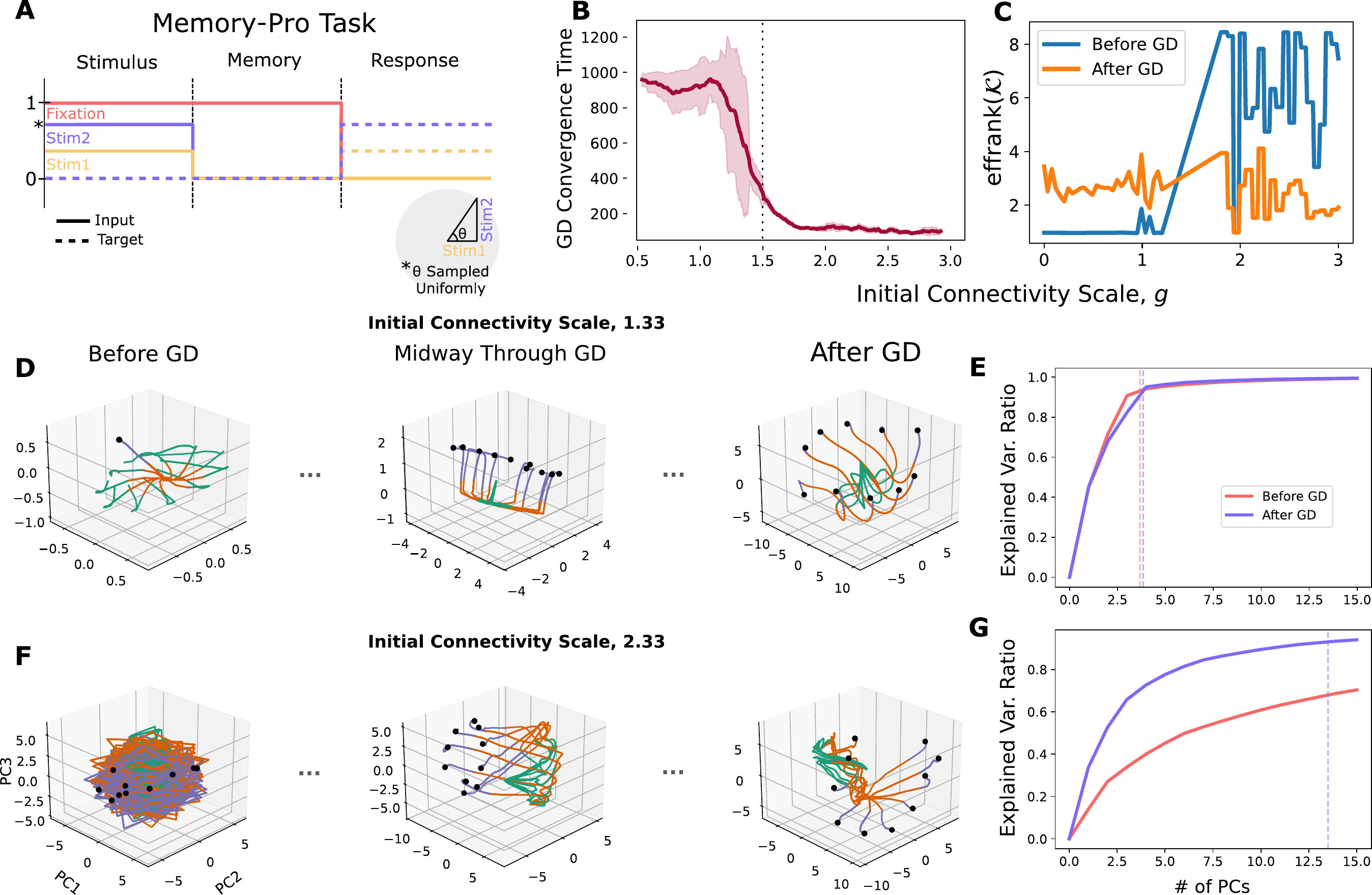}
    \caption{\textbf{Dynamic collapse of model trained on memory-pro task.} \textbf{A} Task inputs and targets, that are piecewise constant over three time periods: \textit{Stimulus} where the network receives an input stimulus, \textit{Memory} where there is no stimulus and the model should not respond, and \textit{Response} where the model should reproduce the same stimuli provided (dashed lines). \textbf{B} Convergence time (number of GD iterations to reach a threshold loss) given initial $W$ connectivity scale, $g$, filtered with a sliding window of length 20, error bars illustrate variance between runs in this window. \textbf{C} Effective rank of $\K$ operator for the models in B. \textbf{D, F} Example hidden state trajectories over three GD iterations (0, 1000 and 5000, respectively) for $g = 1.33$ and $g = 2.33$. Axes are defined by principal components (PCs), computed distinctly for each plot. Each panel shows the hidden state over different task inputs, colored by task period (green-stimulus, orange-memory, purple-response). Black dots denote the final state $z(t_{end})$. \textbf{E, G} Dimensionality of hidden state dynamics before and after training for the two examples, with effective rank (number of PCs required to capture 95 percent variance) indicated. Note that before GD, the $g$ = 2.33 case required around 40 PCs.}
    
    \label{fig:exampledynamics}
\end{figure}



\section{Experiments}

\subsection{Dynamic Collapse on the Memory-Pro Task}

\label{sec:dyncollapse}

\paragraph{Experimental Setup} Prior work has shown that recurrent models such as RNNs and GRUs consistently exhibit collapsed hidden dynamics that form low-dimensional attractors \citep{farrell2023from, sussillo2013opening}. Despite recent advances, especially in simplified models such as linear RNNs \citep{bordelon2025dynamicallylearningintegraterecurrent, NEURIPS2020sch} (also see detailed Related Work in Appendix~\ref{sec:ap1}), we still do not have  a unified theory of attractor formation under GD in non-linear, finite models. 

To investigate how our operator decomposition can contribute, we trained 500 RNNs with varied initial recurrent weight scale, $g$, from $0$ to $3$, where the Lyapunov dimension becomes maximal, with recurrent weights $W_{ij} \sim \mathcal{N}(0, g^2 / n)$. As an illustrative example, the \textit{memory-pro} task was chosen \citep{driscoll2024flexible}, which requires the network to memorize and reproduce two stimuli, drawn as $(x,y)$ coordinates on a a circle, after a memory period in which no input is present (Figure \ref{fig:exampledynamics}A). Consistent with prior findings, models typically converged to a low-dimensional ring of fixed points when projected into PCA space, regardless of $g$ (Figure \ref{fig:exampledynamics}D,F). In Appendix~\ref{sec:apex1}, we reproduced this experiment and operator analysis with ReLU instead of tanh (leading to higher rank $\K$ but still very low relative to $\P$) and GRUs. 

\paragraph{Empirical Observations} Consistent with prior work \citep{engelken2020lyapunovspectrachaoticrecurrent, kadmon2015transition}, we found that as $g$ increases, the dynamics exhibit a transition from collapse to chaos. Remarkably, GD consistently converged faster in the large $g$ regime (Figure \ref{fig:exampledynamics}B). For low $g$, below around $1.5$, the dynamics are initially low dimensional exhibiting collapse to a single fixed point (Figure \ref{fig:exampledynamics}D). With large $g$, the dynamics start out chaotic and higher dimensional (Lyapunov dimension is shown in Appendix~\ref{sec:apex1}). These chaotic models were rapidly damped into a stable regime \citep{liu2024how}. After training, larger $g$ resulted in a higher dimensional hidden state than in the small $g$ networks (Figure \ref{fig:exampledynamics}E,G), but a similar effective rank of $\K$ (Figure \ref{fig:exampledynamics}C). To better understand the mechanisms behind GD collapse on this non-linear model, we turned to the KP operator decomposition. 


\begin{figure}[t] 
    \centering
    \includegraphics[width=.75\linewidth]{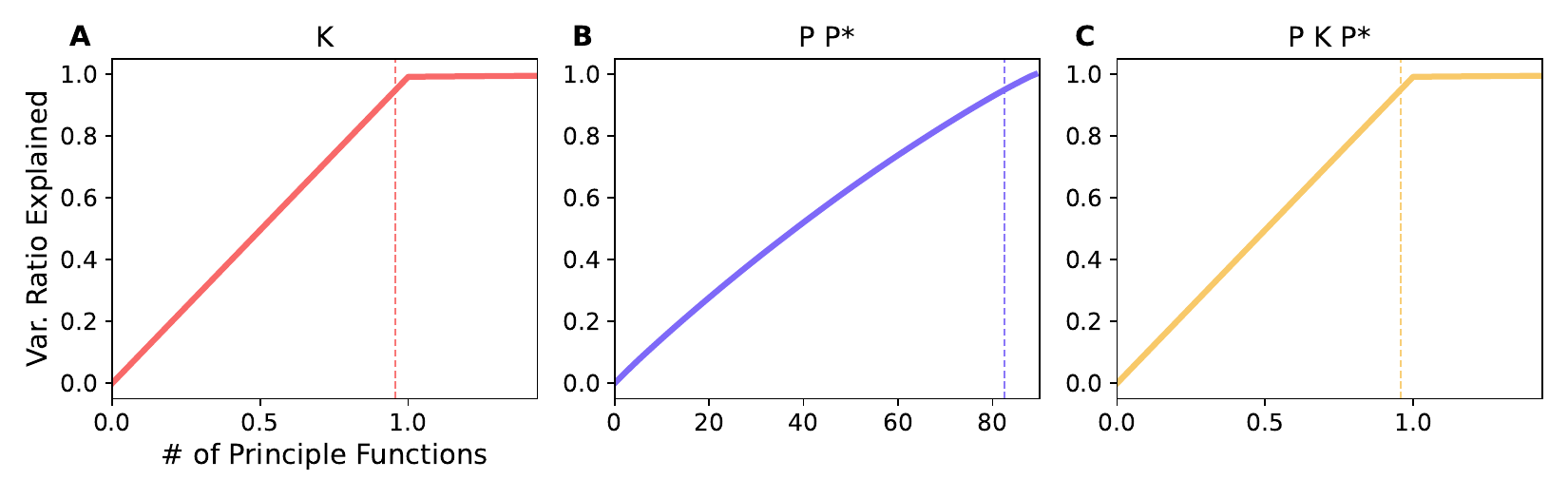}
    \caption{\textbf{$\K$ operator dramatically bottle-necks effective rank of learning.} \textbf{A-C} Cumulative explained variance ratio explained by the eigenfunctions of the operators, $\K$, $\P$ and $\P \K \P^*$, respectively, corresponding to an RNN at initialization with weight scale $g = 1$. The SVD is a consensus over all trial inputs $x$ and hidden units (see Appendix~\ref{sec:consensus}). Note that $\P$ has effectively very high rank, while $\K$, and consequently $\P \K \P^*$, is very low rank.}
    \label{fig:effrankcompare}
\end{figure}

\begin{figure}[t]
  \centering
  \includegraphics[width=.95\linewidth]{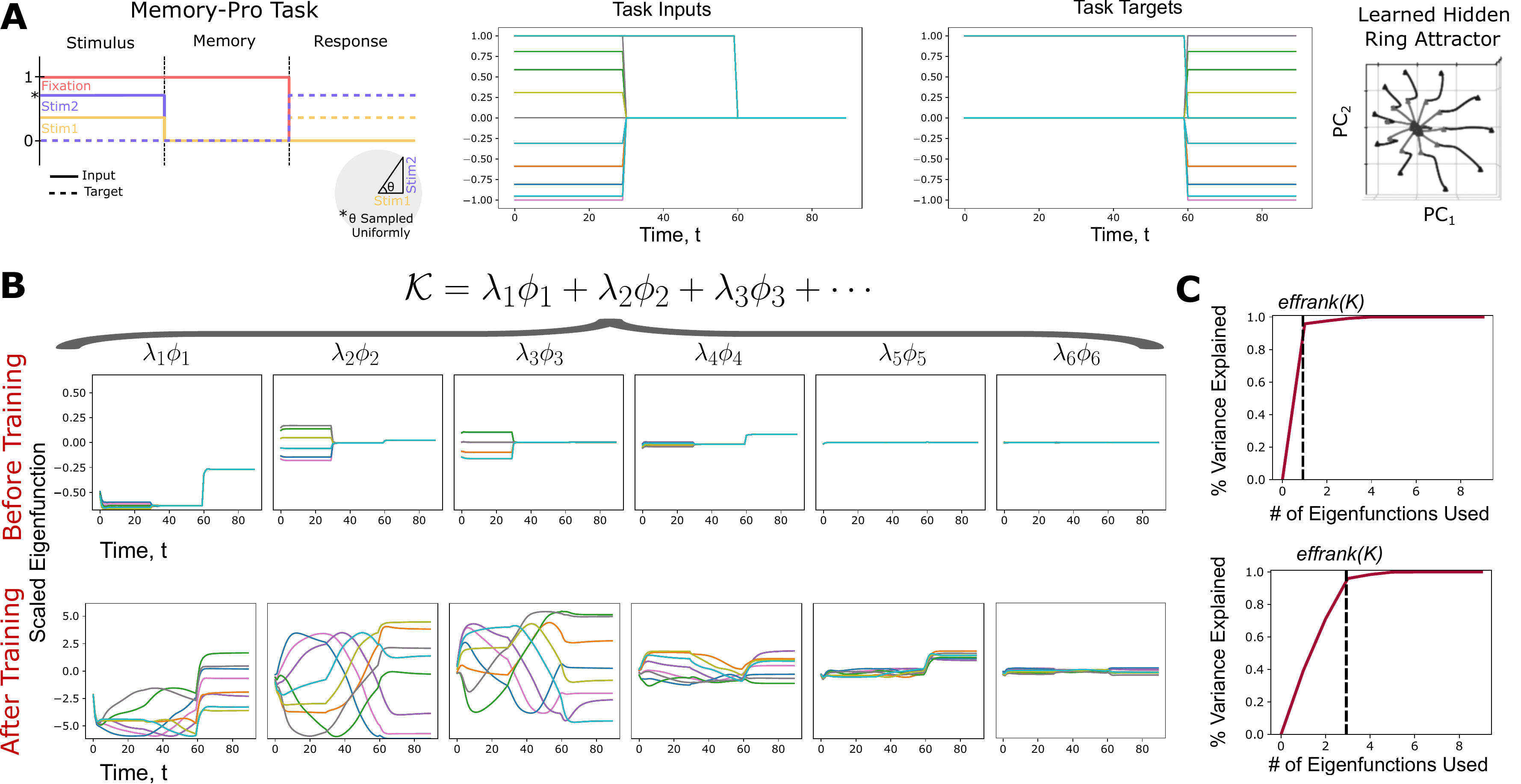}
  \caption{\textbf{Parameter operator $\K$ is low rank for RNN trained with GD on the \textit{memory pro} task.} \textbf{A} Task setup, input and target over every trial $x \sim X$ for the task. Colors correspond to a single choice of $x$ and are consistent in B. \textbf{B} Consensus SVD decomposition of $\K$ measured over all hidden units (see Appendix~\ref{sec:consensus}) for the choice $g = 1.33$ from Figure \ref{fig:exampledynamics}. Before and after training, $\K$ is effectively low rank and explainable by the dominant eigenfunctions. Before training, these directly reflect the task inputs and there are effectively four non-zero dominant modes. After training the eigenfunctions are more complex, corresponding to the learned hidden dynamics, but still the effective rank is low. Finally, \textbf{C} Number of modes of $\K$ needed to explain 95\% of the variance described by its eigenfunctions, i.e. its effective rank.}
  \label{fig:eigfuns}
\end{figure}

\paragraph{Effective Rank of Operators} We measured the reduced SVD of $\K$ and $\P$ as a consensus over all trials, times and computed the effective rank before and after GD (Figure \ref{fig:exampledynamics}C and \ref{fig:eigfuns}). For the RNN, $\K$ has an effective rank bounded by the sum of the effective ranks of $x$ and $\sigma(z)$ due to its simple structure. In particular, we find that $\K$ is typically very low rank; even for $g = 3$ it was only about 7 dimensional, as in panel C. As $g$ increases, so too does the effective rank of $\K$. Crucially, at the end of training, we find that the rank of $\K$ is constrained to a similar range for the given task, regardless of $g$. Similarly, the GRU has very low rank $\K$ in all cases (see Appendix~\ref{sec:apex1}). However, we found that $\P$ is consistently high rank for any $g$ and throughout training (Figure \ref{fig:effrankcompare} and \ref{fig:eigfunsP}), as summarized in the following observation.

\begin{observation}{Choice of Model Directly Constrains Effective Dimension of GD Flow} \\
    For the RNN and GRU, $\K$ is significantly low rank relative to the operator $\P$, regardless of initial recurrent weight scale $g$ throughout GD. Consequently, dynamic collapse appears in such models due to the bottle-necking of dimension by the operator $\K$ in $\P \K \P^*$.
\end{observation}

Section \ref{sec:theory} rigorously motivates observation and Figure \ref{fig:effrankcompare} compares the effective ranks of $\P, \K$ and $\P \K \P^*$ for an RNN at initialization with $g = 1$. Figures \ref{fig:eigfuns} and \ref{fig:eigfunsP} show the breakdown of the eigenfunctions and effective ranks of $\P$ and $\K$ before and after training. Hypothetically, the $\P \P^*$ operator corresponds to the ``parameter-free'' gradient descent, in which the perturbations $\delta f$ are exactly set to the adjoint $a$. Practically implementing learning in this parameter-free way remains elusive, but the operator $\P \P^*$ gives a theoretical baseline to compare $\P \K \P^*$ to for a particular operator $\K$. In Appendix~\ref{sec:paramfree}, we elaborate on this distinction, resembling the difference between \textit{optimize-then-discretize} and \textit{discretize-then-optimize} approaches in optimal control and PDEs \citep{nijmeijer1990nonlinear}. There we explicitly show how the operator $\K$ can misdirect adjoints produced by $\P^*$ and give rise to lower rank learning. 


\paragraph{Summary} Our KPFlow decomposition provides a interpretation of the observed faster training with larger $g$ and dynamical collapse. When $g$ is small, $\text{effrank}(\K)$ is small, hence the error corrections get filtered through an operator that zeros out almost everything. On the other hand, when $g$ is large, $\text{effrank}(\K)$ is larger, leading to less zeroing of the adjoint corrections; furthermore it produces corrections that are more distributed over a range of dimensions. Low rank updates are induced by the bottle-necking of the $\K$ operator in all cases. At the end of training, we find that $\K$ achieves a similar effective rank across different initializations. 


\subsection{Measuring Interference Between Objectives in a Multi-Task Context}

\label{sec:interferencemulti}

\begin{figure}[t]
    \centering
    \includegraphics[width=.925\linewidth]{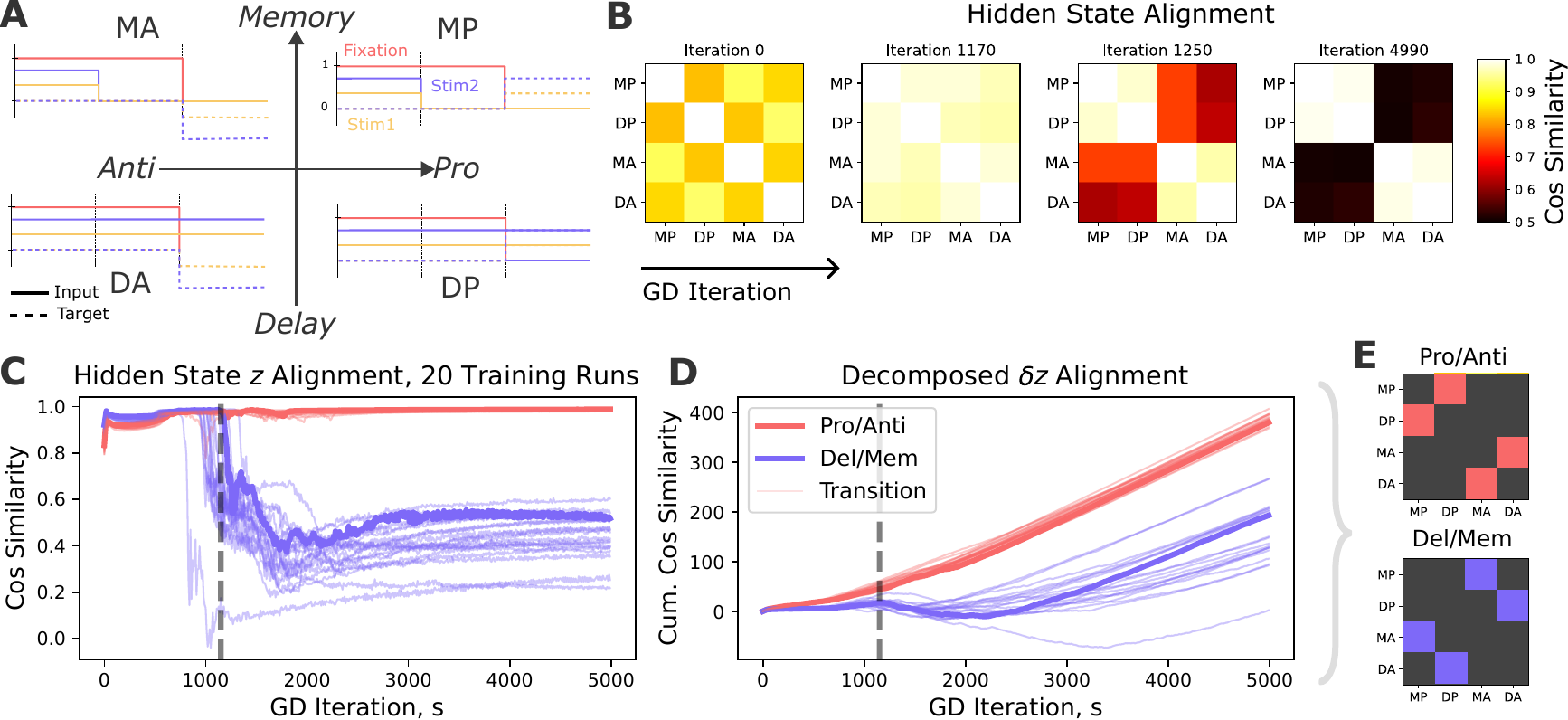}
    \caption{\textbf{Interference matrix determines emergence of subspace-aligned, shared dynamics among four tasks.} \textbf{A} Schematic of inputs and targets for four related tasks Memory Anti (MA), Memory Pro (MP), Delay Anti (DA), and Delay Pro (DP), on which 20 GRU networks were simultaneously trained. \textbf{B} Cosine similarity matrices over training, measuring alignment between all hidden state trajectories for each sub-task with consistent inputs. Note that after iteration 1250, the hidden states organize so that hidden unit activations for MP and DP most strongly align, as well as MA and DA. \textbf{C} Normalized sum of cosine similarity for tasks with the same pro/anti orientation (red) or tasks with the same delay/memory requirement (blue), showing that all 20 training runs exhibited a consistent transition away from delay/memory alignment around the same iteration. Bold lines correspond to run shown in B. \textbf{D} Cumulative alignment based on the interference matrix (see main text), showing that there is a preference towards pro/anti alignment throughout GD, which is not readily visible prior to the transition in B and C. \textbf{E} Filters used on matrices in B to produce C and D, where colored cells equal 1, black cells are 0.}
    \label{fig:multi}
\end{figure} 

\paragraph{Decomposing the KPFlow for multiple Tasks} Here, we show how the KPFlow further decomposes in a multi-task example where $X$ is composed of two disjoint tasks $X_1, X_2$. Then, with the notation of \ref{sec:prelim}, any collection of trajectories over trials, $q \in \T$, can be written in the form $q = [q_1, q_2]^T$, where $q_j$ denotes the value of $q(t|x)$ restricted to all trials $x \sim X_j$ from task $j$. Then, $\P$ and $\K$ take block-matrix form $\K = \begin{bmatrix} \K_{11} & \K_{12} \\ \K_{21} & \K_{22} \end{bmatrix}, \, \, \P = \begin{bmatrix} \P_1 & 0 \\ 0 & \P_2 \end{bmatrix}$, where $\P_j$ is the $\P$ restricted to trials from $X_j$, $\K_{ij}$ translates error signals from task $j$ into corrections restricted to task $i$. In particular,
\begin{align}
    \label{eqn:breakdown}
    \delta z = \begin{bmatrix} \delta z_1 \\ \delta z_2 \end{bmatrix} = \begin{bmatrix} \P_1 \K_{11} \P_1 & \P_1 \K_{12} \P_2 \\ \P_2 \K_{21} \P_1 & \P_2 \K_{22} \P_2 \end{bmatrix} \begin{bmatrix} \Err_1 \\ \Err_2 \end{bmatrix}.
\end{align}
where $\Err_j$ is the error signal from task $j$ trajectories. The operators $\P_i \K_{ij} \P_j$ measure how error signals on trial $j$ interfere with dynamics on trial $i$. Consequently, we can write $\delta z_1 = \delta z_{11} + \delta z_{12}$, where $\delta z_{11}$ is a correction to $z_1$ directly from $\Err_1$, while $\delta z_{12}$ represents interference from $\Err_2$. To quantify how these terms align, i.e. how the interference aligns with sub-task corrections, we define an \textbf{interference matrix}, $M$, given by the cosine similarities $M_{ij} = \langle \delta z_{ii} , \delta z_{ij} \rangle / (\|\delta z_{ii}\| \| \delta z_{ij}\|)$.

\paragraph{Experimental Setup} As a second illustration, we consider multi-task neural networks, trained on four tasks simultaneously. The tasks were Memory-Pro (MP), as above (Figure \ref{fig:exampledynamics}); Memory-Anti (MA), in which the target is in the reverse direction of the stimulus; Delay-Pro (DP), where there is no memory period and the stimulus is present throughout the task; and Delay-Anti (DA), with no memory period and reversed target directions. See Figure \ref{fig:multi}A. The network is provided a context input over an initial additional duration for each trial which informs the network of the specific sub-task. After this duration, the context input is turned off, instead of being on for the whole trial, as in prior work \citep{driscoll2024flexible}. 

\paragraph{Understanding shared representations via the interference matrix}


We trained 20 GRU models the four tasks with ADAM and initial weight scale $g = 1$. Consistent with prior work \citep{driscoll2024flexible,turner2023simplicity,NEURIPS2020sch,lea2020}, networks reused rings of fixed points that were aligned across similar tasks. To quantify this, we measured the cosine-similarity over all times and hidden units between the hidden states over trials restricted to the four tasks, $z_i$ and $z_j$ for $i, j = 1 \cdots 4$, producing four by four task alignment matrices at each GD iteration (Figure \ref{fig:multi}B). For each task, the stimuli used to compare similarities between tasks were identically set. From the produced matrices, we applied two filters to measure preferential alignment between tasks based on two metrics: aligning tasks based Pro/Anti or based on Memory/Delay (Figure \ref{fig:multi}E). Note that sharing based on cosine similarity implies well-aligned activity subspaces, but does not necessarily address other notions of sharing, such as re-use of fixed points across attractors, as in prior work \citep{driscoll2024flexible}. 

Remarkably, a consistent transition to aligning tasks based only on pro/anti occurred around a similar GD iteration (Figure \ref{fig:multi}C). This transition was not readily visible in the cosine similarity plots prior to it occurring. However, we found that it was possible to anticipate this transition via the interference matrix described above,  We first compute $M(s)$, accumulated over GD iterations $s$ (filtering it in the same way as in panel C) leading to Figure \ref{fig:multi}D. This term measures how well the individual interfering components of $z$ are aligned with the desired corrections, accumulated as GD constructs the individual sub-task trajectories. We found in all trials that this metric suggested that alignment based on pro/anti was promoted from the start of, and throughout, task learning. Thus, the interference matrix revealed the information relevant to the eventual sharing of dynamics among the four tasks. 

More figures and an additional metric based on Rayleigh quotients of the interference operators $\P_i \K_{ij} \P_j^*$ appear in Appendix~\ref{sec:apex2}. Furthermore in Appendix~\ref{sec:apex3}, we provide an extension of Experiment merging Experiment 1, in which we swept the GRU's initial recurrent weight scale, $g$, between $0$ and $10$, analyzing the emergence of alignment between sub-tasks in this context for all cases, finding that similar separation based on Pro/Anti consistently occurred for $g < 6$ and more overall alignment with less predictable structure appeared with $g > 6$. We hypothesize that with small $g
$, similar tasks reuse the same attractors because their perturbations during gradient descent align. With large $g
$, the system becomes chaotic and task alignment depends on navigating the complex loss landscape rather than task similarity. Both regimes need further investigation. Overall, we view the present work as opening doors toward more detailed investigation into the structure of these operators, how they induce aligned attractors, building a richer theory of attractor re-use in multi-task neural networks \citep{driscoll2024flexible, turner2023simplicity}. 

\section{Theoretical Results} 

\label{sec:theory}



In this section, we provide an analysis of the properties of $\P$ and $\K$, supporting our observations above. The following Theorem describes the properties of $\K$ for a typical class of models in machine learning and neuroscience \citep{rumelhart1986learning, gerstner2014neuronal} (as a diverse example, we derive the KPFlow for a Hodgkin-Huxley biophysical neural network in Appendix~\ref{sec:hhappend}). 

\begin{theorem}
    \label{thm:operator}
    Suppose the model in Equation \ref{eqn:forwarddyn} is \textit{weight-based}, $\theta = \{W_1, ..., W_M\}$, with each $W_j$ applied once in a single evaluation of $f$. Then, (1) $\mathcal{K}$ is a sum of $M$ operators induced by each weight, $\K = \sum_{j=1}^M \K_j$. (2) Each $\mathcal{K}_j$ is a positive semi-definite Hilbert-Schmidt integral operator induced. (3) The effective rank of $\K_j$ over time and trials is bounded by the effective dimension of the dynamical quantity to which $W_j$ is applied.
\end{theorem}

See Appendix~\ref{sec:appendixproofop} for a proof. Property 3 is most relevant to our Experiments, motivating why $\mathcal{K}$ is low rank for the models considered, relating this rank to the effective dimension of the model's dynamics. As above, for the RNN in Equation \ref{eqn:rnn}, $\mathcal{K} = \K_1 + \K_{2}$, corresponding to the weights $W_{in}, W$; the rank is bounded by summing the effective dimensions of $\sigma(z(t|x))$ and $x(t)$ over all $t$ and $x$. Next, the following Theorem decomposes $\P$ into a product of operators. 
%
%

\begin{figure}[t]
  \centering
  \includegraphics[width=.91\linewidth]{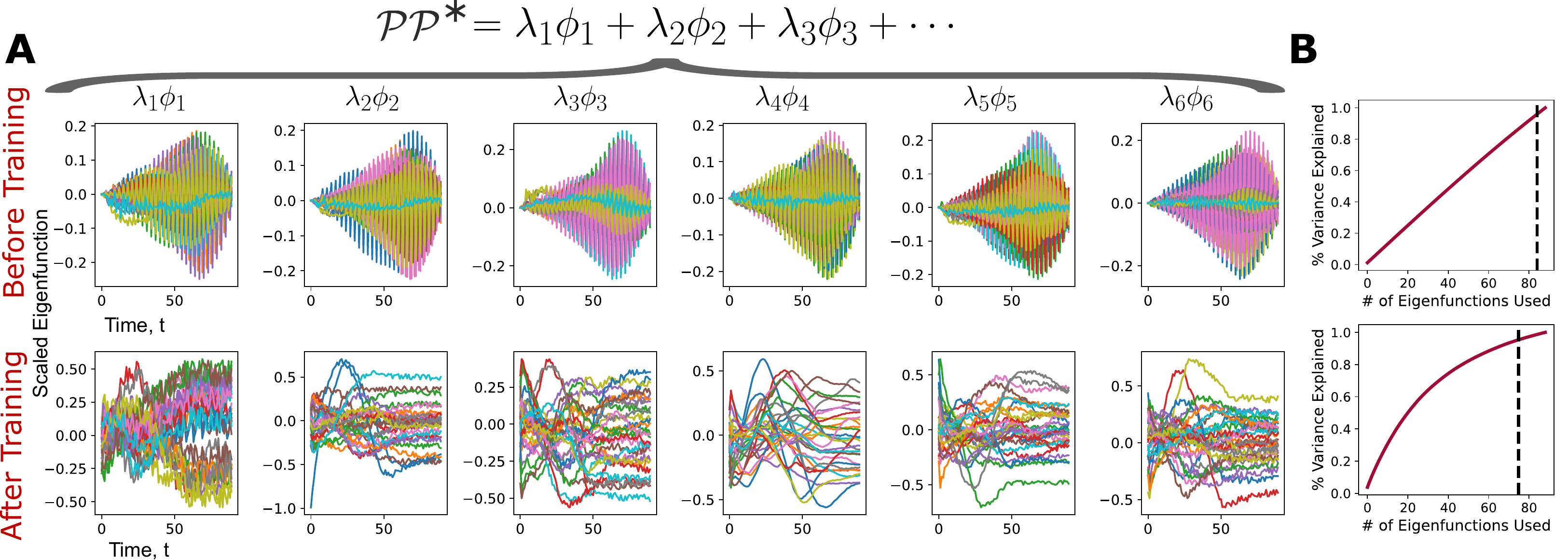}
  \caption{\textbf{$\P$ operator is high rank before and after GD on the memory-pro task.} Similar to Figure \ref{fig:eigfuns}, \textbf{A} eigenfunctions and \textbf{B} effective rank of $\P$ before and after training for an example training run where $g = 1.33$. The eigenfunctions of $\P$ exhibit oscillation and the singular values are evenly distributed. At the end of training, there is more structure, but the effective rank is still high.} 
  \label{fig:eigfunsP}
\end{figure}

\begin{theorem}
    The operator $\P$ decomposes as $\P = U V U^{-1}$ where (1) $U$ is the ``fundamental operator'' for the dynamics, propagating changes to the initial conditions of the model to perturbations to the hidden state at time $t$, (2) $V$ is the Volterra operator, $(Vq)(t|x) = \int_0^t q(t_0|x) \d t_0$. The trial- and time-dependent full singular values of $U$ are $\sigma_j(t|x) = \exp(\lambda_j(x) t)$ where $\lambda_j(x)$ is the $j$th Lyapunov exponents of the system on trial $x \sim X$. $V$ singular values that scale as $O(1/j)$. 
\end{theorem}

The proof is in Appendix~\ref{sec:thm2proof} Example eigenfunctions and SVD of $\P$ are shown in Figure \ref{fig:eigfunsP}.  We observe empirically that $\P$ is high rank in all Experiments. The operator $\P \P^*$ used to compute the SVD of $\P$ generalizes the controllability Grammian in optimal control\citep{nijmeijer1990nonlinear}. Future work could leverage theory in this area to more explicitly derive the structure and SVD of $\P$ in our context.

\vspace{-.4cm}

\section{Discussion} 
\vspace{-.2cm}

In this work, we introduce \emph{KPFlow}, a decomposition of the GD flow associated with training of a dynamical system in terms of two operators, $\K$ and $\P$, acting on the space of tensors over time, task trials and hidden units. Implementing these operators and their SVDs reveals a trove of structure in each stage of GD. The explicit breakdown of the gradient flow into $\P$ and $\K$ is natural since $\P$ encodes linear time propagation and is observed to be much higher rank for the models considered, while $\K$ constrains GD by filtering through parameter changes, making it much lower rank and inducing dynamic collapse. In addition, only $\K$ mixes between trials, facilitating further breakdown of $\P \K \P^*$ into operators in a multi-task context that each describe how sub-task corrections align.
\vspace{-.2cm}
\paragraph{Limitations and opportunities.} KPFlow sets the stage for future work addressing the limitations or our current work. These limitations include the need and opportunities to rigorously derive why $\P$ is high rank, and to consider specialized models, such as those with unitary Jacobians or infinite width limits. Furthermore, while we have identified the low-rank structure of $\K$, we have not yet applied it to inform new learning rules. Future research could also investigate how $\P$ and $\K$ simplify under assumptions on the dynamics, e.g., under fast hidden state collapse to fixed points $\P$ may simplify dramatically. Finally, the tasks considered here are relatively simple, chosen for illustration. Considering more diverse tasks could be addressed in future work.

\textbf{Our code} is available at \url{https://github.com/meeree/adjoint_dynamics}.

\bibliographystyle{unsrt}
\bibliography{refs}

\clearpage
\appendix

\subfile{appendix.tex}

\section*{NeurIPS 2025 Checklist}

\begin{enumerate}

\item {\bf Claims}
    \item[] Question: Do the main claims made in the abstract and introduction accurately reflect the paper's contributions and scope?
    \item[] Answer: \answerYes{} 
    \item[] Justification: Our claims are re-stated in the contributions at the end of the introduction section and outlined in the Experiments and Theoretical Results sections.

\item {\bf Limitations}
    \item[] Question: Does the paper discuss the limitations of the work performed by the authors?
    \item[] Answer: \answerYes{} 
    \item[] Justification: Our work's limitations and future directions are provided in the Discussion section.
    
\item {\bf Theory assumptions and proofs}
    \item[] Question: For each theoretical result, does the paper provide the full set of assumptions and a complete (and correct) proof?
    \item[] Answer: \answerYes{} 
    \item[] Justification: Yes, proofs are provided in the Appendix and informal observations are motivated with experimental findings, noting in these cases that the observations are not explicitly proven. 

    \item {\bf Experimental result reproducibility}
    \item[] Question: Does the paper fully disclose all the information needed to reproduce the main experimental results of the paper to the extent that it affects the main claims and/or conclusions of the paper (regardless of whether the code and data are provided or not)?
    \item[] Answer: \answerYes{} 
    \item[] Justification: Code with Jupyter notebooks reproducing the plots are provided and the Appendix and main text detail the implementation.

\item {\bf Open access to data and code}
    \item[] Question: Does the paper provide open access to the data and code, with sufficient instructions to faithfully reproduce the main experimental results, as described in supplemental material?
    \item[] Answer: \answerYes{} 
    \item[] Justification: Yes, our code is linked in the main text and it contains python notebooks reproducing the experiments.

\item {\bf Experimental setting/details}
    \item[] Question: Does the paper specify all the training and test details (e.g., data splits, hyperparameters, how they were chosen, type of optimizer, etc.) necessary to understand the results?
    \item[] Answer: \answerYes{} 
    \item[] Justification: Details are provided in experimental setup and in the Appendix. 

\item {\bf Experiment statistical significance}
    \item[] Question: Does the paper report error bars suitably and correctly defined or other appropriate information about the statistical significance of the experiments?
    \item[] Answer: \answerYes{} 
    \item[] Justification: Error bars are included in plots where relevant and multiple runs are performed of all experiments. 

\item {\bf Experiments compute resources}
    \item[] Question: For each experiment, does the paper provide sufficient information on the computer resources (type of compute workers, memory, time of execution) needed to reproduce the experiments?
    \item[] Answer: \answerYes{} 
    \item[] Justification: Details of computer setup are provided in the Appendix; a standard workstation was used for all experiments. 
    
\item {\bf Code of ethics}
    \item[] Question: Does the research conducted in the paper conform, in every respect, with the NeurIPS Code of Ethics \url{https://neurips.cc/public/EthicsGuidelines}?
    \item[] Answer: \answerYes{} 
    \item[] Justification: The research does conform.

\item {\bf Broader impacts}
    \item[] Question: Does the paper discuss both potential positive societal impacts and negative societal impacts of the work performed?
    \item[] Answer: \answerYes{} 
    \item[] Justification: Broad discussion of our work to machine learning is provided in the Discussion section.
    
\item {\bf Safeguards}
    \item[] Question: Does the paper describe safeguards that have been put in place for responsible release of data or models that have a high risk for misuse (e.g., pretrained language models, image generators, or scraped datasets)?
    \item[] Answer: \answerNA{} 
    \item[] Justification: We do not introduce new models or datasets.

\item {\bf Licenses for existing assets}
    \item[] Question: Are the creators or original owners of assets (e.g., code, data, models), used in the paper, properly credited and are the license and terms of use explicitly mentioned and properly respected?
    \item[] Answer: \answerYes{} 
    \item[] Justification: The paper authors created all code and experiments and generated data in the paper.  

\item {\bf New assets}
    \item[] Question: Are new assets introduced in the paper well documented and is the documentation provided alongside the assets?
    \item[] Answer: \answerYes{} 
    \item[] Justification: The Appendix and main text provide a comprehensive summary of introduced code and methods.

\item {\bf Crowdsourcing and research with human subjects}
    \item[] Question: For crowdsourcing experiments and research with human subjects, does the paper include the full text of instructions given to participants and screenshots, if applicable, as well as details about compensation (if any)? 
    \item[] Answer: \answerNA{} 
    \item[] Justification: We did not use crowdsourcing or human subjects.

\item {\bf Institutional review board (IRB) approvals or equivalent for research with human subjects}
    \item[] Question: Does the paper describe potential risks incurred by study participants, whether such risks were disclosed to the subjects, and whether Institutional Review Board (IRB) approvals (or an equivalent approval/review based on the requirements of your country or institution) were obtained?
    \item[] Answer: \answerNA{} 
    \item[] Justification: We did not use human subjects.

\item {\bf Declaration of LLM usage}
    \item[] Question: Does the paper describe the usage of LLMs if it is an important, original, or non-standard component of the core methods in this research? Note that if the LLM is used only for writing, editing, or formatting purposes and does not impact the core methodology, scientific rigorousness, or originality of the research, declaration is not required.
    \item[] Answer: \answerNA{} 
    \item[] Justification: LLMs are not a core component of our research and they were not used in generation of the paper.

\end{enumerate}

\end{document}

%% file: appendix.tex
\begin{center}
{\centering \LARGE \textbf{Appendix to KPFlow}}
\end{center}

\begingroup
   \small
   \setlength{\parskip}{0pt}
   \setlength{\parindent}{0pt}
   \tableofcontents
\endgroup


\section{Expanded Related Work}

\label{sec:ap1}

\captionsetup[figure]{labelfont={bf,it}, name=Supplementary Figure}


\paragraph{Theoretical Foundations of Gradient Dynamics in Neural Networks}

\subparagraph{Neural Tangent Kernel} As detailed in the main paper, our work is related to the NTK in that it studies the gradient flow dynamics through the lens of operators. However, it differs in multiple ways from the classical and current NTK literature. Classically, the NTK was proposed to explain the exact gradient flow of feed-forward neural networks \citep{jacot2018neural}. It was shown that in the infinite width limit, the kernel matrix associated with this gradient flow becomes constant. Subsequent work extended this to the RNN architecture \citep{alemohammad2020recurrent}. Recent work extended the NTK further to certain Neural ODEs by formulating it as an infinite limit of the NTKs associated with ResNet neural networks \citep{gao2025exploring}. KPFlow differs in several aspects from these works. In particular, the primary difference is that we propose the \textit{specific KP decomposition}, organizing the general operator associated with learning for \textit{any} recurrent dynamical system into a product involving two operators. We show that this organization is logical and can be used to understand how the individual steps of GD can cause dynamical collapse and sharing of attractors in multi-task contexts. 
Furthermore, the theoretical background describing the operators as acting on a space of 3-tensors differs. Thus, our work provides a specific decomposition and generalizes the gradient flow analysis to any recurrent model. Our experimental focus and emphasis on empirical use of the operators as tools for analyzing emergence of recurrent dynamics also differs from prior work. The operators we describe in our main work act on the space of all ``per-trial trajectories'' which, when discretized, are tensors of the form [B, T, H] with batch dimension B, time dimension T and hidden count H. The eigenfunctions corresponding to the SVD of these operators are thus also of shape [B, T, H], specifying a trove of information about how exactly to maximally stimulate each operator involved in backpropagation, leading to fast learning, at every individual instant in forward-pass time, $t$, batch trial, $x$, and hidden unit. 


\subparagraph{Linear Case} Prior work has investigated similar phenomena to us experimentally and theoretically through the lens of linearized RNNs. For example, the recent elegant work of \citep{bordelon2025dynamicallylearningintegraterecurrent} explicitly proves how GD trains a linear RNN on an integration task. In our case, these is not a general closed form way of integrating the gradient flow over GD iterations due to the model being nonlinear. Instead, we use the operators defined in the main text as empirical and theoretical analysis tools with properties that are stated in the main text. Similar linearized analysis occurs in \citep{NEURIPS2020sch}. Prior work has also shown with mean field models of infinite width feed-forward neural networks how the GD flow evolves, breaking learning into multiple stages \cite{mei2018mean}. Our work generalizes the analyses to more general recurrent dynamical systems, proposing a specific decomposition of the gradient flow associated operators that could help to theoretically extend the analysis of prior work to more complex models. 


\paragraph{Dynamical Analysis of Recurrent Dynamics}

\subparagraph{Foundational Works}

Our work primarily analyzes GD learning on sequential models by utilizing tools from dynamical systems. From this viewpoint, the dynamics of a model are interpreted through fixed points, attractors, periodicity, and through asymptotic and local measures of stability \citep{sussillo2013opening, mante2013context, tang2022open, farrell2023from}. This perspective found a strong foothold in theoretical neuroscience \citep{izhikevich2007dynamical, gerstner2014neuronal, dayan2001theoretical} and, more recently, throughout literature interpreting learned representations of recurrent neural networks (RNNs) and related models. Sussillo and Barak were among the first to demonstrate that recurrent models learn structured, interpretable representations through gradient decent made up of fixed points and attractors on a variety of tasks \citep{sussillo2013opening}. The tasks presented in this work have been intensively studied in other works. For example, the ``3-bit-flip-flop'' task requires a network to store an internal state and transition between states based on inputs. It is is observed in multiple works that recurrent models solve this task through a ``cube attractor'' with fixed points at vertices of the cube \citep{jarne2024exploring, sussillo2013opening}. In our work, we use this task as an illustrative example throughout. 

Outside of trained representations, much classical work has analyzed the dynamics of RNNs with weights sampled from $\mathcal{N}(0, g / \sqrt{n})$, with hidden dimension $n$, when $g$ is scaled \citep{sompolinsky1988chaos, rajan2006eigenvalue, sussillo2009generating}. There, it is observed that networks exhibit a transition from collapse to fixed points with small weight scale to chaos with large weight scale.

%
%
\subparagraph{Simplicity Bias, Compositional Learned Representations and In-Domain Transfer Learning}

Much work has used the dynamical systems perspective on learned representations to frame hypotheses and observations about how RNNs learn to solve tasks. It is observed that such models exhibit a so-called ``simplicity bias'' (or the related notion of compression of learned representations, or ``neural collapse'') preferentially converging to simple attractor structures on single tasks, and re-using attractors and subspaces when trained on multiple distinct tasks \citep{turner2023simplicity, farrell2023from, driscoll2024flexible, Farrell2022, Papyan2020, RecanatesiFarrell2021, Tishby1999}. Dynamical motifs (e.g. learned attractor structures) can be more robust to new unseen inputs and noise, supporting better generalization \citep{sussillo2013opening}. Moreover, further work showed that neural networks can learn representations that are compositional for multiple distinct but related tasks \citep{yang2019task}. In particular, networks build the full task dynamics as a composition of sub-tasks, supporting better generalization to new tasks composed of these sub-tasks \citep{driscoll2024flexible}. In transfer learning, it has been observed that trained dynamical systems models generalize better to tasks that are ``in domain/manifold'' \citep{goring2024out}. In our work, we provide new theoretical tools to address these phenomena. In particular, we aim to show that tools such as the adjoint dynamical system and parameter kernels could provide insights into this GD flow, possibly allowing for a better explanation of how single- and multi-task simplicity bias manifests.

\subparagraph{Lyapunov Exponents for RNNs} 

One specific theoretical tool that has seen more recent adoption as a means of interpreting learning in RNNs are Lyapunov exponents. Unlike local stability analysis at a fixed point, these exponents describe the long-term stability of a model along trajectories in the infinite time limit \citep{vogt2022lyapunov}. By Oseledet's theorem, the Lyapunov spectrum is provably invariant under the choice of trajectory under certain ergodicity assumptions on the model \citep{ruelle1979ergodic}. Finite time Lyapunov exponents, as their name suggests, describe the stability of a dynamical system under finite timeframes for a chosen trajectory \citep{storm2024finite}. Recent work has shown that the Lyapunov spectra can address questions related to generalization capabilities and expressiveness of RNNs \citep{mikhaeil2022difficulty, goring2024out}. In our work, we extend these results through the operator $\P$ and its implementation. We show below (\ref{sec:oppractical}) how to compute the operator in a way similar to algorithms computing the Lyapunov exponents. The operator $\P$ produces the perturbed state $\delta z$ given changes to the tangential dynamics on all trial inputs $x$ and at every timestep $t$. Classical Lyapunov exponents describe the change $\delta z$ to the state after perturbations to the initial conditions of $z$. Thus, the operator $\P$ captures a more complex, learning-relevant metric where the entire trajectory is perturbed, not just the initial condition. How the SVD of $\P$ relates to the Lyapunov spectra exactly is an area for future research. Theorem 2 provides a first step in this direction.


\paragraph{Optimal Control and the Adjoint Method}

\subparagraph{Neural ODEs} 


Early work by LeCun et al. demonstrated that the gradient backpropagation algorithm is mathematically equivalent to the adjoint-state method from optimal control theory \cite{lecun1988theoretical}. In this framework, the network's forward pass is interpreted as a dynamical system, and the backward pass corresponds to integrating an associated costate system as prescribed by Pontryagin’s maximum principle. In particular, backpropagation through time (BPTT) \citep{rumelhart1986learning} for recurrent neural networks (RNNs) can be viewed as applying the adjoint method to the unfolded state dynamics, computing the influence of earlier states on the final loss via a backward recursion. This optimal control perspective provides a theoretical foundation for understanding learning in sequential models and deep networks as trajectory optimization problems for the network’s state evolution.


\subparagraph{Optimal Control Perspective on Learning} The dynamical systems viewpoint has further catalyzed theoretical insights into deep learning by casting training as a gradient flow or optimal control problem in continuous time. Weinan E \cite{weinan2017proposal} proposed viewing supervised learning as steering a dynamical system to approximate a target function, treating network parameters as time-varying control variables. Similarly, Li et al. \cite{li2017maximum} translated Pontryagin’s optimality conditions into novel training algorithms that extend beyond standard gradient descent. These works reinforce the equivalence between backpropagation and the backward integration of the adjoint equations, thereby establishing a principled approach  for exploring the evolution of network dynamics during training.

%
%
\begin{figure}
    \centering
    \includegraphics[width=.8\linewidth]{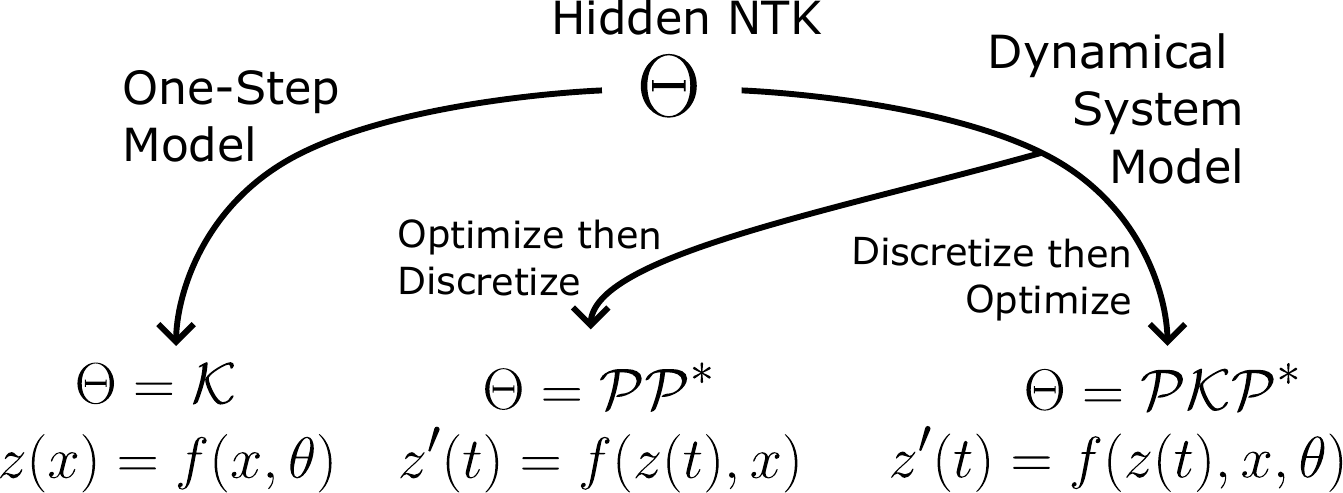}
    \caption{\textbf{The operator $\Theta$, as in Corollary 1 of the main text, related to the hidden state gradient flow takes specialized forms based on the presence of time-stepping and the optimization approach.} When the model has no time-stepping, i.e. $z(x) = f(x | \theta)$, e.g. a feed-forward neural network, $\Theta$ is just $\K$. When the model is instead a time-varying dynamical system, $z'(t|x) = f(z(t|x), x(t) | \theta)$, the model can be trained in two ways. Firstly, GD can directly carry the adjoint to perturbations to $f$, in which case $\Theta = \P \P^*$. Since the gradients are not filtered through parameters, this is known as the \textit{optimize-then-discretize} methodology \citep{onken2020discretize}. Alternatively, as is typical in deep learning, the changes to $\delta f$ are produced by filtering the adjoint through parameters, leading to $\Theta = \P \K \P^*$, as in the paper, the so-called \textit{discretize-then-optimize} approach. Note that the case where neither $\K$ or $\P$ appears, i.e. $\Theta = \Id$ corresponds to linear regression each GD step. }
    \label{fig:ntkcases}
\end{figure}
%
%
%
%

%
%

\section{Theoretical Background}

\label{sec:ap2}

\subsection{GD Flow Derivation}
\label{sec:ap21}

\paragraph{Supervised Learning Context}

We let $s$ denote the Gradient Descent (GD) iteration. In the simplest form of GD, the parameters $\theta(s)$ evolve as $s$ progresses according to 
\begin{align}
    \theta_{s+1} = -\eta \nabla_{\theta(s)} L(\theta(s)).
\end{align}
Once the trial input $x(t) \sim X$ is chosen, there is assumed to be a corresponding target for this particular trial input, $y^*(t|x)$. Then, the loss (which implicitly depends on $\theta$ through $z$) is assumed to have the form
\begin{align}
    L(\theta) &:= \E_{x \sim X} [\int_0^{t_{end}} \ell(z(t | x,\theta), y^*(t|x)) \d t],
\end{align}
where the term $\ell$ defines the instantaneous ``running loss'' on each individual timestep. Throughout our work, $\ell$ is an MSE loss with linear readout with dynamically trained $W_{out}, b_{out}$ and normalization constant $k$ defined by the batch size, timesteps and number of hidden units,
\begin{align}
    \ell(z(t|x), y^*(t)) = \frac{1}{k} \| W_{out} z(t|x) + b_{out} - y^*(t)\|. 
\end{align}

\paragraph{Adjoint Method Preliminaries}

Fundamentally, GD necessitates the ability to compute efficient and accurate per-trial loss gradients, $\nabla_\theta  L(\theta |d)$. The most common means of computing such gradients in deep-learning contexts is backwards-mode automatic differentiation (also known as backpropagation through time, BPTT), which is provably identical to the \textit{adjoint method} in both continuous and discrete cases \citep{lecun1988theoretical,pontryagin1962mathematical}. The latter is distinguished as a more general approach to solving ODE- and PDE-constrained optimization problems primarily in the context of optimal control. We describe the general detials of this in the main text, but here we go into more depth.

As in Supplementary Figure \ref{fig:adjointschematic}, the adjoint method computes per-trial gradients by solving three initial value problems \citep{chen2018neural}: 
\begin{enumerate}
    \item \textit{Forward propagation}, where the state $z(t|x)$ is simulated from time $0$ to time $t_{end}$. 
    \item \textit{Adjoint backpropagation}, where an adjoint variable is simulated backwards from time $t_{end}$ to time $0$.
    \item \textit{Gradient backpropagation}, where a running gradient is simulated backwards from time $t_{end}$ to time $0$ effectively ``extrapolating'' the adjoint in state space to parameter space sensitivities. 
\end{enumerate}

Step 1 is as detailed above. Step 2 computes the per-trial adjoint, $a(t | d)$, defined as
\begin{align}
    \label{eqn:adjoint}
    a(t | d) := \nabla_{z(t|d)} L(\theta | d).
\end{align}
Empirically, the adjoint measures how small perturbations tangential to the hidden state trajectory at a an instant in time $t$ (i.e. in the direction of $f$) affect the loss. The adjoint is computed incrementally backwards as below. 

First, we define 
\begin{align}
    \Err(t |d) := \nabla_{z(t|x)} \ell (z(t|x), y^*(t)),
\end{align}
which is an instantaneous sensitivity of the running loss with respect to state. The notation $\Err(t | d)$ is motivated by the special case of a regression task, where 
\begin{align}
    \label{eqn:mse}
    \ell(z(t|x), y^*(t)) = \| W_{out} z(t|x) - y^*(t)\| \,\, \Rightarrow \, \, \Err(t | d) = 2 W_{out}^T (W_{out} z(t|x) - y^*(t)),
\end{align}
i.e. when using a squared-error loss $\Err(t | d)$ is simply an extrapolated residual. 

The dynamics of the adjoint are given by the following initial value problem \citep{chen2018neural}:
\begin{align}
    \frac{\d}{\d t} a(t | d) &= - D_{z(t | x)} f(t|d)^T \cdot a(t | d) - \Err(t | d); \, \,  a(t_{end} | d) = 0. 
\end{align}
The term $D_{z(t|x)} f$ denotes the Jacobian of $f$ at $z(t|x)$. The forcing term,


Step (3) defines a running gradient sensitivity $g_\theta(t | d)$ in the parameter space, which is typically higher dimensional than the hidden state space. Explicitly, it is given by 
\begin{align}
    \label{eqn:subgrad}
    g_\theta(t | d) &= \nabla_{z(t|d)} \int_t^{t_{end}} \ell(z(t_0|x, \theta), y^*(t_0)) \d t_0. 
\end{align}
The gradient $g_\theta(t |d)$ is given by solving backwards the initial value problem \citep{chen2018neural}:
\begin{align}
    \label{eqn:runninggrad}
    \frac{\d}{\d t} g_\theta(t | d) = - D_\theta f(t|d)^T \cdot a(t|d); \, \, g_\theta(t_{end} | d) = 0.  
\end{align} 
Finally, from Equation \ref{eqn:subgrad}, the per-trial loss gradient is defined by
\begin{align}
    \nabla_\theta  L(\theta |d) = g_\theta(0 | d).
\end{align}
Finally, given all trials, $d$, the proposed parameter update is a mean of these parameter proposals: 
\begin{align}
     \delta  \theta  = -\eta \underset{d \sim D}{\E} [ g_{\theta(s)}(0 | d) ]. 
\end{align}

\begin{figure}[t]
    \centering
    \includegraphics[width=0.5\linewidth]{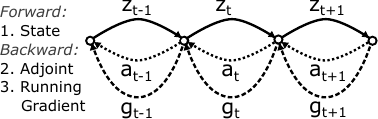}
    \caption{\textbf{Schematic of the Discretized Adjoint Method for Sequential Models}. }
    \label{fig:adjointschematic}
\end{figure}

\paragraph{Discrete Case}

We briefly describe the discrete analogue of these equations, which are exactly the equations of BPTT \citep{lecun1988theoretical}. In this case, the model has the form
\begin{align}
    z_{t+1 | x} = \mathfrak{f}(z_{t|x} | x, \theta),
\end{align}
and the loss is a sum instead of an integral. The adjoint has recurrent dynamics given by 
\begin{align}
    a_{t | y^*} = D_{z_{t+1|x}} \mathfrak{f}^T \cdot a_{t+1|y^*} + \nabla_{z_{t|x}} \ell(z_{t|x}, y^*_t), \, \, \, a_{t_{end} | y^*} = 0
\end{align}
and the running gradient is given by 
\begin{align}
    g_{\theta,t | y^*} = D_\theta \mathfrak{f}(z_{t | x} | x, \theta)^T \cdot a_{t+1 | y^*}.
\end{align}


\paragraph{GD Flow}

We first define the linearized flow, relevant to the operator $\P$ in the main text. Define the \textit{state-transition matrix} on trial $x$, $\Phi(t, t_0 | x)$, by 
\begin{align}
    \label{eqn:statetransmat}
    \frac{\d}{\d t} \Phi(t, t_0 | x) &=  J_z(t|x) \cdot \Phi(t, t_0 | x) \text{ for } t > t_0, \,\, \, \, U(t_0, t_0 | x) = \text{Id}, 
\end{align}
This matrix defines how instantaneous tangential perturbations $\delta f(t_0 | x)$ translate to changes $\delta z(t|x)$ at time $t$ to the hidden state. The finite and long time Lyapunov spectra and directions, which quantify behavior of the dynamical trajectories $z(t|x)$ as a growth or decay in certain directions, can be directly derived from $\Phi$ \citep{vogt2022lyapunov}.

%

Using the state-transition matrix, we can then write the full gradient flow as 
\begin{align}
\label{eqn:full}
\delta z(t | x) &= -\int_0^t \underbrace{\Phi(t,t_0 | x)}_{{\color[HTML]{323031} \substack{\text{Forward} \\ \text{Propagator}}}} \, \underset{\substack{(x_1, y_1^*) \\ \sim D}}{\mathbb{E}} \Big[
    \int_0^{t_{\text{end}}} 
    \underbrace{D_\theta f(t_0 | x) D_\theta f(t_1 | x_1)^T}_{{\color[HTML]{1B9AAA} \text{Parameter Kernel}}}
    ( \int_{t_1}^{t_{\text{end}}} \underbrace{\Phi(t_2,t_1 | x_1)^T}_{{\color[HTML]{323031} \substack{\text{Backward} \\ \text{Propagator}}}} \, \underbrace{\Err(t_2 | y_1^*)}_{{\color[HTML]{DB2955} \text{Correction}}} \d t_2 )
    \d t_1 
\Big] \d t_0
\end{align}

This is now proven:

\begin{proof}
At a time $t_0$ and trial $x$, the perturbation to the dynamics, $f$, can be computed through the linear term of the Taylor expansion for reasonably small weight updates:
\begin{align}
    \delta f(t_0 | x) &= -D_\theta f(t_0 | x) \cdot \delta \theta \\
    &= -D_\theta f(t_0 | x) \cdot \underset{\substack{(x_1, y_1^*)\\ \sim D}}{\E}[g_\theta(0)] \\
    &= -D_\theta f(t_0 |x)  \underset{\substack{(x_1, y_1^*)\\ \sim D}}{\E} [\int_0^{t_{end}} D_\theta f(t_1 | x_1)^T \cdot a(t_1 | y_1^*) \d t_1] \\
    &= -\underset{\substack{(x_1, y_1^*)\\ \sim D}}{\E} [\int_0^{t_{end}} D_\theta f(t_0 |x)   D_\theta f(t_1 | x_1)^T \cdot  (\int_{t_1}^{t_{end}} \Phi(t_2, t_1 | d_1)^T \Err(t_2 | y_1^*) \d t_2 )\d t_1].
\end{align}
So, the full formula follows by plugging the above into
\begin{align}
    \delta z(t|x) &= \int_0^t \Phi(t, t_0 | x) \delta f(t_0 | x) \d t_0.
\end{align}
\end{proof}

\subsection{Formal Operator Definition}

\label{sec:ap22}

\paragraph{$\T$ Space}
The gradient flow \ref{eqn:full} can be written in terms of operators, as in our main work. First, we define the space $\T$ of per-trial trajectories in hidden space formally. For a single trial input $x$, the space of trajectories $\T_x$ with $x$ as the trial input is given by $L^2([0, t_{end}]; \R^n)$, where $n$ is the number of hidden units. In general, $\T$ is a direct sum of $\T_x$ over all trials, i.e. if there are $B$ trial inputs, we define $\T$ to be a space of $B$ copies of $L^2([0, t_{end}]; \R^n)$. If $X$ is an infinite measurable space, we formally define $\T$ to be a ``direct integral'' written as $\T = \int_X^\oplus \T_x \d x$, meaning that sampling a trial $x \sim X$ specifies a space of trajectories, $\T_x$, conditioned on that trial choice. 

\paragraph{$\P$ Definition}
With $\T$ and the state-transition matrix $\Phi$ given, we define the \emph{Propagation Operator}, $\mathcal{P} : \mathbb{T} \rightarrow \mathbb{T}$ by 
\begin{align}
    \label{eqn:statetransitiondef}
    [\mathcal{P} q](t | x) &=\int_0^t \Phi(t_0, t | x) q(t_0) \d t_0,
\end{align}
for any $q \in \T$, a per-trial trajectory.

\textit{Action:} The operator $\P$ can be described in terms of its action. In particular, given a change, $\delta f$, to the forcing term, $\d z / \d t = f$, $\P \delta f = \delta z$ gives the corresponding change to $z$. So, we can evaluate $\P \delta z$ by running a model twice, once with the original forcing term $f$, and once with the new forcing term, $f + \delta f$. In reality, this only works for reasonably small perturbations since $\P$ is defined through linearization, so $\P$ can be thought of as the linear part of the transformation $\delta f \mapsto \delta z$. 

\paragraph{$\K$ Operator} Define the \emph{Parameter Operator}, $\mathcal{K}  : \mathbb{T} \rightarrow \mathbb{T}$ by
\begin{align}
    [\mathcal{K} q](t | x) = J_\theta (t|x) \underset{x \sim X}{\E}\Big[\int_0^{t_{end}} J_\theta(t_0|x_0)^T  q(t_0 | x_0) \, \d t_0 \Big].
\end{align}
The expected value and inner integral produces a single static quantity, $\delta \theta$, in parameter space. $J_\theta(t |x)$ maps this perturbations to the parameters to a change to $z(t|x)$. 


\subsubsection{Discrete Informal Definition}

\label{sec:ap221}

In practice, even with neural ODEs \citep{chen2018neural}, time and batch trials are discrete when forward-evaluating the model. For simplicity, we assume here that the trial-dependent time interval, $[0, t_{end}]$, is the same for all trials. We let $\tilde{}$ denote the discretized version of an object. Suppose we sample a batch $\tilde X$ of length $B$ from $X$, evaluate the model at $T$ times $0 = t_1 \leq \cdots \leq t_T = t_{end}$ and there are $n$ neurons. Then, 
\begin{align}
    \tilde \T &= \{\text{arrays of shape } [B, T, n]\}.
\end{align}
Consequently, discretized operators on this space map tensors of shape $[B, T, n]$ to tensors of the same shape, i.e. they can be viewed as very large matrices, 
\begin{align}
    \tilde \K, \tilde \P \in \R^{(B\cdot T \cdot n) \times (B \cdot T \cdot n)}. 
\end{align}

\paragraph{Parameter Operator}
The matrix $\tilde \K$ is dense in the general context, but it behaves identically as a scalar multiple of the identity for some models, such as the RNN. Explicitly, indexing $\tilde \K$ with $b, b_0 \leq B, t, t_0 \leq T$ and using $:$ notation,
\begin{align}
    \tilde \K[b, t, :, b_0, t_0, :] = \tilde J_\theta[b,t] \tilde J_\theta[b_0, t_0]^T.
\end{align}
If the parameter dimension of $\theta$ is $m$, $J_\theta[b, t]$ is an $n$ by $m$ matrix giving the Jacobian with respect to $\theta$ at discrete time $t$ and only batch trial $b$. 

\paragraph{Propagation Operator} There are two alternative ways to define the propagation operator, which both agree for small inputs, where perturbations to the hidden state are well described by the linearized flow (see Implementation section in this Appendix \ref{sec:oppractical}). We briefly describe the discretized linearized form. Assume for simplicity that we have a discrete map, $z_{n+1} = \mathfrak{f}(z_n, x_n)$ and let $\tilde J_z[b, t]$ be the Jacobian corresponding to $\mathfrak{f}$ on batch $b$ and time $t$. Let $\tilde p$ denote the output of $\tilde \P \tilde q$ applied to an input signal $\tilde q$ which is a tensor in $[B, T, n]$. Then, $\tilde p$ is given by solving
\begin{align}
    \tilde p[b, t+1, :] = \tilde J_z[b, t] \cdot  \tilde p[b, t, :] +  \tilde q[b, t+1, :],
\end{align}
with $\tilde p[b, 1, :] = q[b, 1, :]$. In other words, we iteratively multiply the Jacobian and add $\tilde q[b, t+1, :]$, on each individual batch $b$ separately.

%

Since the discretized operators are extremely large, it is impossible to compute the entire matrix representation in most cases (for example, if $n = 1,000$, $B = 100$ and $T = 100$, a generic operator would require $10^{14}$ entries to be stored). However, computing the action of the operators is cheap in memory and time, from which most of the useful properties such as the eigenvalues and eigenfunctions can be computed. Indeed, there are many existing tools providing linear algebra functionality in this large operator case \citep{golub2013matrix,kreyszig1989introductory,2020SciPy-NMeth}. 

\subsection{Consensus Reduction of Operators} 

\label{sec:consensus}

In this section, we describe an approach for reducing the operators involved in this work. These operator act on per-trial trajectories which are generally 3-tensors over all times, $t$, task trials and hidden units. SVD can still be applied to such operators, but in many cases it makes more sense to measure a ``reduced SVD'' as a consensus over certain axes, e.g., taking a consensus over all task trials and times, which specifies directions in hidden space, verged over all time and trials. First, we describe how to measure a consensus Lyapunov spectrum for the specific fundamental operator $U$, as in Theorem 2 of the main text. Following this, the same approach is extended to any general operator. The methods involved are implemented in our \texttt{KPFlow} software, supporting computing an \texttt{AveragedOperator} version of any given operator over any particular axes.

\paragraph{Consensus Lyapunov spectra example}

For a single trial, let $\sigma_i(t|x)$ be the finite time singular values of $U(t|x)$. Then, the finite time Lyapunov exponents of the dynamical system with fixed trial input $x$ are  
\begin{align}
    \lambda_i(t|x) &= \frac{1}{t} \cdot \log(\sigma_i(t|x)). 
\end{align}
Taking the final value of $t$ gives the best ``long time'' value for this Lyapunov exponent. As $t \rightarrow \infty$, $\lambda_i(t|x)$ stably approaches the true Lyapunov exponent for almost any trajectory \citep{ruelle1979ergodic}. However, there are still some technical details needed to quantify the ``shared Lyapunov spectrum'' of the ensemble of trajectories over all inputs $x(t)$. 

When discretized over trials, the fundamental matrix looks like a block matrix over all trials $x_1, x_2, ..., x_B$:
\begin{align}
    U(t) &= \begin{pmatrix} U(t|x_1) & 0 & 0 & \dots \\
    0 & U(t | x_2) & 0 & \dots \\
    0 & 0 & U(t | x_3) & \dots \\
    \vdots & \vdots & \vdots & \ddots \end{pmatrix}.
\end{align} 
So, taking the SVD of this matrix will act block-wise and there are Lyapunov exponents for each individual trial. To form a consensus, instead we propose to form the matrix,
\begin{align}
    U_{wide}(t) &= \frac{1}{B} \begin{pmatrix} U(t | x_1) & U(t|x_2) & \dots \end{pmatrix},
\end{align}
assuming there are $B$ trials. The SVD of this matrix naturally accounts for subspace sharing between multiple trajectories. Note that finding the SVD of $U_{wide}$ is equivalent to taking the eigendecomposition of the ``averaged controllability Grammian matrix'' matrix
\begin{align}
    G_{consensus}(t) &:= \E_{x} [U(t|x) U(t|x)^T],
\end{align}
so practically we can compute the averaged matrix above and take its eigendecomposition to obtain Lyapunov exponents capturing variability over all trials. 

\paragraph{Reduction of Any Operator} Consider instead a general operator, $\mathcal{W}: \T \rightarrow \T$, e.g., the parameter operator $\K$ or the propagation operator $\P$. Suppose we want to take a consensus over trials and times. Then, we define an ``averaged operator'' $\langle \mathcal{W} \rangle_{x,t} $ acting solely on the space of hidden units, $\R^n$. For any input $\hat q \in \R^n$, we define an element $q \in \T$ in the full 3-tensor space given by $q(t|x) = \hat q$, i.e. we just identically copy the input over all trials and times. Then, we define $\langle \mathcal{W} \rangle_{x,t} \cdot \hat q := \E_{x, t} [W q]$, i.e. the averaged operator applies the action of $W$ to an identically duplicated copy of the input over all trials and times, then averages the result, which is a tensor, over all trials and times. Conceptually, taking the SVD of this operator agrees with the consensus SVD described above, i.e. the SVD of $\langle W \rangle_{x,t}$ is the same as forming the averaged Grammian $G^{x,t}_{consensus} = \E_{x,t}[\mathcal{W} \mathcal{W}^T]$, as in the Lyapunov example, and performing and eigendecomposition. The axes we choose to average over can arbitrary. We implement these methods in our \texttt{KPFlow} code through the \texttt{AveragedOperator} class, which wraps any given operator, facilitating taking a consensus over any given axes. 

\subsection{Parameterized Learning and Parameter-Free Learning}

\label{sec:paramfree}

The operator $\P \K \P^*$ transferring errors to the gradient flow represents learning relative to a choice of parameters $\theta$ that are used to train the model. Alternatively, we can let the adjoint directly perturb the flow without filtering through parameters $\theta$, i.e. $\delta f = a$. Learning with $\P \P^*$ is optimal since it allows for the most direct changes to the flow, $f$, as below, but in reality it is not desirable (or practically possible) since perturbing along the adjoint exactly produces a model that is out of the parameter regime we desire. 

\paragraph{Gradient Misdirection} Write $\P = U \Sigma V^T$, so that $\P \P^* = U \Sigma^2 U^T$. Then, 
\begin{align}
    -\P \P^* \Err = -\sum_i \sigma_i^2 u_i \cdot \langle \Err, u_i \rangle,
\end{align}
i.e. error in a ``direction'' $u_i$ will translate perfectly to a correction in the $u_i$ direction, scaled by the variance of that direction. When we instead learn with the operator $\P \K \P^*$, we get ``gradient misdirection,'' since 
\begin{align}
    \P \K \P^* &= U (\Sigma V^T \K V \Sigma) U^T,
\end{align}
and the internal operator $B:=\Sigma V^T \K V \Sigma$ is typically not the identity, as in the parameter-free case with no $\K$ operator. Note in this case that 
\begin{align}
    -\P \K \P^* \Err = -\sum_i b_i \cdot \langle \Err, u_i \rangle,
\end{align}
and $b_i$ may not necessarily agree with $\sigma_i^2 u_i$. Thus, there is a ``gradient misdirection'' based on the alignment between $\K$ and $\P$. This can be formally quantified by checking how well aligned the principle directions of $U B$ and $U \Sigma^2$ are. Thankfully, if $U$ is effectively low rank, $k$, $B$ and $\Sigma$ are $k$ by $k$ and we can measure the low rank projection alignment between $\Sigma^2$ and $B$. Note that $B$ is positive definite, so $B = A^2$ for some matrix $A$, and hence it's equivalent to check that $A$ aligns with $\Sigma$, but only up to sign. 


\section{Implementation Details}

\label{sec:oppractical}

\paragraph{Parameter Operator} Applying the parameter operator $\K q$ can be broken into two steps. First, we compute
\begin{align}
    \theta_q := \underset{x_0 \sim X}{\E} \Big[ \int_0^{t_{end}} J_\theta(t_0 | x_0) q(t_0 | x_0) \d t_0 \Big].
\end{align}
This is a single static quantity in the parameter space. Then, for each trial $x$ and time $t$, we produce the output $(\K q)(t|x)$ by the Jacobian vector product
\begin{align}
    (\K q)(t|x) &= J_\theta(t|x) \theta_q.
\end{align}
This step is efficiently implemented in Pytorch \citep{paszke2019pytorchimperativestylehighperformance} (and JAX \citep{jax2018github}) as the \texttt{jvp} operation, which never explicitly forms the Jacobian $J_\theta$. On the other hand, the first step above, computing $\theta_q$ is given by the $\texttt{vjp}$ operation. In particular, we can easily and efficiently compute $\K q$ by first applying a $\texttt{vjp}$ then applying $\texttt{jvp}$ to the result. 

\paragraph{Propagator Operator} There are multiple ways to compute the propagation operator. We implement two ways: 
\begin{enumerate}
    \item \textit{The linear approach}, using the state-transition matrix definition in Equation \ref{eqn:statetransitiondef},
    \item \textit{The direct approach}, relying on perturbing the model dynamics and measuring how the hidden state is perturbed. 
\end{enumerate}

\paragraph{The linear approach} relies on the variation of parameters for small inputs $q$. The linearized form of $\P$ produces perturbations $\delta z$ given changes $\delta f$ to the tangential dynamics, assuming that these are reasonably small (i.e. linearization of the operator applies)/ In particular, if $p$ is the output of $\P q$ for $q \in \T$ that is small in norm on all trials, then $p$ satisfies the ODE 
\begin{align}
    \frac{\d}{\d t} p(t|x) = J_z(t|x) \cdot p(t|x) + q(t|x), \, \, \, p(0|x) =0.
\end{align}
We found that solving this ODE gives an effective means of computing the operator $\P$ for our applications. In the discrete case, it just consists of iteratively multiplying by $J_z(t|x)$ for all times $t \in [0, t_{end}]$ on all trials $x \sim X$. We also implemented a more stable version based on QR approaches to computing the state-transition matrix, $\Phi(t, t_0 | x)$, extending prior work focused on computing Lyapunov exponents for recurrent models \citep{vogt2022lyapunov}, which are derived from the singular values of $\Phi$. In the discrete case, for each trial input $x$, we let $Q_0^x = \text{Id}$ and we compute over all times
\begin{enumerate}
     \item  $M_{t+1}^x = J_z(t|x) \cdot Q_t^x$,
     \item $Q_{t+1}^x, R_{t+1}^x = $\texttt{QR}$(M_{t+1}^x)$,
\end{enumerate}
Then, we compute $\Phi(t, t_0 | x) = Q_t^x R_t^x R_{t-1}^x \cdots R_{t_0+1}^x Q_{t_0}^x$ for $t > t_0$ and $\Phi(t, t | x) = \text{Id}$. Finally, the value of $(\P q)(t|x)$ is given by the state-transition solution Equation \ref{eqn:statetransitiondef}. This approach is likely more stable since it accounts for very small or large Jacobian values by applying QR at every iteration \citep{vogt2022lyapunov}, but in reality we found that directly solving the variational dynamics above by iteratively multiplying the Jacobians directly was sufficient in our case. 

\paragraph{The direct approach} The second way to compute $\P q$ is by directly perturbing the model dynamics, which we found to be generally slightly faster than the former linear approach, as well as more memory efficient since it doesn't require computing Jacobian matrices. In particular, if $\delta f$ is a small change to the forcing dynamics, $f$, of the model, the 
\begin{align}
    \P : \delta f \mapsto \delta z 
\end{align}
i.e. $\P$ maps the changes at every timestep to the term $\delta f$ to the change $\delta z$ to the model hidden dynamics, integrated over times. Hence, we can compute 
\begin{align}
    \P q \approx \text{forwards}(f + q) - \text{forwards}(f),
\end{align}
where forwards simulates the model with a given right hand side, $f$, over all trials and time. This is not guaranteed to be an operator in general since it may not even be linear, but for small $\delta f$ it agrees with the linear form above. We provide code to validate that the two approaches produce the same output to very small relative error for small $\delta f$. Note that the form above does require being able to simulate the model forwards with the perturbed $f + q$ dynamics, which may not always be possible without a user explicitly implementing this change to their model. In this case, the linear version can be used which does not rely on simulating forwards the perturbed dynamics.

\paragraph{Computing SVDs} We implement the operators above as an instance of a generic \texttt{Operator} class. This class can be converted into a scipy \citep{2020SciPy-NMeth} \texttt{LinearOperator}, which supports operator SVD and other decompositions. These are implemented entirely in a \textit{matrix-free} way, i.e. all that is needed is the action of the operator and its Hermitian adjoint, we do not need to form the full discretized operator matrix. We also implement a class \texttt{AveragedOperator}, which assumes inputs are reduced over some axes (e.g. trials, times or hidden units, or multiple at once) to measure the consensus SVD, reducing over these axes, as outlined in Section \ref{sec:consensus}.
\section{Proofs}

\label{sec:proofs}

\subsection{Proof of Proposition 1}

The detailed derivation from which the proof follows is in \ref{sec:ap21} above. 

\label{sec:appendixproofflow}

\begin{proof}
    At a time $t_0$ and trial $x$, the perturbation to the dynamics, $f$, can be computed through the linear term of the Taylor expansion for reasonably small weight updates:
    \begin{align}
        \delta f(t_0 | x) &= -D_\theta f(t_0 | x) \cdot \delta \theta \\
        &= -D_\theta f(t_0 | x) \cdot \underset{\substack{(x_1, y_1^*)\\ \sim D}}{\E}[g_\theta(0)] \\
        &= -D_\theta f(t_0 |x)  \underset{\substack{(x_1, y_1^*)\\ \sim D}}{\E} [\int_0^{t_{end}} D_\theta f(t_1 | x_1)^T \cdot a(t_1 | y_1^*) \d t_1] \\
        &= -\underset{\substack{(x_1, y_1^*)\\ \sim D}}{\E} [\int_0^{t_{end}} D_\theta f(t_0 |x)   D_\theta f(t_1 | x_1)^T \cdot  (\int_{t_1}^{t_{end}} \Phi(t_2, t_1 | d_1)^T \Err(t_2 | y_1^*) \d t_2 )\d t_1].
    \end{align}
    So, the full formula follows by plugging the above into
    \begin{align}
        \delta z(t|x) &= \int_0^t \Phi(t, t_0 | x) \delta f(t_0 | x) \d t_0.
    \end{align}
\end{proof}

\subsection{Proof of Corollary 1}

\label{sec:corrproof}

\begin{proof}
    We know that $\delta z = - \P \K \P^* \Err$ by Proposition 1, proven above. In the case of a squared error loss (MSE loss), the error term is equal to 
    \begin{align}
        \Err &= \frac{1}{2 k} W_{out}^T (y - y^*),
    \end{align}
    for some normalization constant dependent given by the product of the number of timesteps, batch size and number of hidden units, assuming we take a mean over all these axes in the loss. We assume for simplicity in the corollary that $2 k = 1$. Note since $y = W_{out} z + b_{out}$, $\delta y = W_{out} \delta z$, so
    \begin{align}
        \delta y = W_{out} \P \K \P^* W_{out}^T (y - y^*), 
    \end{align}
    as desired.
\end{proof}

\paragraph{Dynamic output weights} if the term $W_{out}$ is also trained, then this leads to a change to the error signal, $\Err$, which depends on it, but not directly to the operator $\P$ and $\K$, which are defined internally through the hidden state. In particular, GD consists of two steps. First, we compute $\delta z = \P \K \P^* \Err$ as above. Then, we update the weights $W_{out}$ and the biases. GD on the weights simply performs linear regression to maximize alignment between $y$ and $y^*$ as best is possible. Thus, when $W_{out}$ and $b_{out}$ are trained, the gradient flow breaks into two coupled dynamical updates, repeated one after the other.   

\subsection{Proof of Theorem 1}

\label{sec:appendixproofop}
\begin{proof}
Consider a per-trial trajectory $g \in \mathbb{T}$. The operator is explicitly defined as 
\begin{align}
    (\mathcal{K} g)(t | x) &= \int_D \int_{t=0}^{t_{end}}  D_\theta f(t | x) D_\theta f (t_0 | x_0)^T g(t|x) \d t \, \d \mu(x_0),
\end{align}
The outer product defines the similarity matrix
\begin{align}
    k(t, x, t_0, x_0) = D_\theta f(t | x) D_\theta f (t_0 | x_0)^T \in \R^{n \times n}.
\end{align}


To define the weight-based sequential model with weights $\{W_1, ..., W_M\}$, we suppose that $f(z | x)$ can be computed by multiple ``views'' of the computation graph. For simplicity, we assume \textit{ each weight is only used once in one evaluation of $f$ at an instant in time}, as is the case in most typical models (e.g. LSTMs, GRUs, RNNs, transformers do not ``re-feed'' the signal back into the weights for a single step forwards in time). It's clear from these assumptions that, for each weight index $j$, we can decompose $f$ into a ``view'' of the computation graph: 
\begin{align}
    \label{eqn:views}
    \underbrace{f(z |x) = f^{outer}_j(W_j \cdot f_j^{inner}(z | x), f_j^{other}(z|x))}_{\text{$j$th weight's view of $f$}}. 
\end{align}
This is illustrated in We can decompose the parameter space into a direct sum of each weight's individual parameter spaces, so $D_\theta f$ decomposes as a direct sum, 
\begin{align}
    D_\theta f  &= (D_{W_1} f, ..., D_{W_M} f). 
\end{align}
Let $q_j = W_j f_j^{inner}(z|x)$ in the equation \ref{eqn:views}. Then,


\begin{figure}[t]
    \centering
    \includegraphics[width=0.5\linewidth]{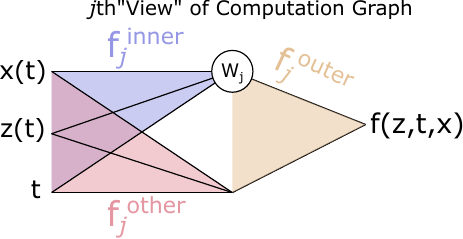}
    \caption{\textbf{Visual depiction of computation graph ``view'' corresponding to a weight $W_j$ involved in computing $f$}. Assuming $W_j$ is only used once in computing $f(z(t|x), t |x)$, we can break the graph into three distinct functions, as shown.}
    \label{fig:view}
\end{figure}

\begin{align}
    k(t,x,t_0,x_0) &= D_\theta f(t | x)  \cdot D_\theta f^T(t_0 | x_0) = \sum_{j=1}^M D_{W_j} f(t | x) \cdot D_{W_j} f^T(t_0 | x_0) \\
    &= \sum_{j=1}^M \langle f_j^{inner}(z(t)|x),  f_j^{inner}(z(t_0)|x_0) \rangle \cdot D_{q_j} f_j^{outer}(t | x) D_{q_j} f_j^{outer}(t_0 | x_0)^T,
\end{align}
where $\langle \cdot \rangle$ denotes the vector inner product. Letting
\begin{align}
    k_j(t, x, t_0, x_0) := \langle f_j^{inner}(z(t|x)),  f_j^{inner}(z(t_0|x_0)) \rangle \cdot D_{q_j} f_j^{outer}(t | x) D_{q_j} f_j^{outer}(t_0 | x_0)^T,
\end{align}
we can thus write, 
\begin{align}
    k(t, x, t_0, x_0) &= \text{Id}_{n \times n} \cdot \sum_{j=1}^M k_j(t, x, t_0, x_0),
\end{align}
i.e. $k$ decomposes into a sum of inner products defined by each view corresponding to the weight parameters. 

In particular, if we define the operator $\mathcal{K}_j : \T \rightarrow \T$ acting on per-trial trajectories, $q \in \T$, by 
\begin{align}
    (\mathcal{K}_j q)(t | x) &= \E_{x_0 \sim X} [\int_{t=0}^{t_{end}}  k_j(t, x, t_0, x_0) q(t_0|x_0) \d t]
\end{align}
then
\begin{align}
    \mathcal{K} = \sum_{j=1}^M \mathcal{K}_j. 
\end{align}
Showing property (1) of Theorem 1. Each operator, $\mathcal{K}_j$, is a kernel-induced Hilbert-Schmidt integral operator. It is thus compact, continuous, self-adjoint and positive definite since it is defined in terms of Grammian matrices and inner products. Since $\mathcal{K}$ is a sum of these operators, it inherits all these properties.

Finally, we consider the effective dimension of $\mathcal{K}_j$. Define an operator induced simply by the inner product term:
\begin{align}
   ( \hat {\mathcal{K}_j} q)(t|x) &:= \E_{x_0 \sim X} [\int_{t=0}^{t_{end}} \langle f_j^{inner}(z(t|x)), f_j^{inner}(z(t_0|x_0)) \rangle \cdot  q(t_0|x_0) \d t].
\end{align}
This operator weights by the autocorrelation between the term $f_j^{inner}(z(t|x))$, which is the dynamical term to which $W_j$ is applied, over all times $t$ and trials $x$. This is a Grammian operator, i.e. when discretized it behaves like matrix-vector multiplication by a matrix Gram matrix of $f_j^{inner}(z(t|x))$, flattened over all time, trials and hidden units. The effective dimension of the trajectory $f_j^{inner}(z(t|x))$ to which $W_j$ is applied is defined in terms of the eigenvalues of this Gram matrix in an identical way to the effective rank of $ \hat {\mathcal{K}_j}$. Thus, the effective rank of this operator, taken as a consensus over hidden units (see \ref{sec:consensus} for a formal definition) is identical to the effective dimension of $f_j^{inner}(z(t|x))$ over all $t$ and $x$. 

Finally, the true operator corresponding to the model in question is $\K_j$, which has also has a recurrent Jacobian outer product term, not $\hat {\K_j}$, which only involves the scalar inner product kernel. In the case of the RNN considered in the main text or the Hodgkin-Huxley network below, both operators are equal for all $j$ weight indices. In general, the operator $\K_j$ has a rank that depends on both $f_j^{inner}$ and $f_j^{outer}$. However, since its kernel is multiplied by the same scalar inner product kernel defining $\hat {\K_j}$, its overall rank is bounded above by the effective dimension of $f_j^{inner}$, instead of being exactly equal. Explicitly evaluating the rank of $\K_j$ itself in terms of $f_j^{outer}$ as well, instead of just bounding it, is an area for potential future investigation. 

However, taken together, we demonstrate that the rank of $\K_j$ is given by the effective dimension of the dynamical quantity, $f_j^{inner}$, to which $W_j$ is applied, being bounded by this dimension when $f_j^{outer}$ is not the identity, and equal to this dimension when it is, e.g., in the case of the RNN or Hodgkin-Huxley biophysical network below.
%
%
%
%
\end{proof}

\subsection{Proof of Theorem 2}
\label{sec:thm2proof}

\begin{proof} 
    Note the fundamental operator $U$ is defined by 
    \begin{align}
        (U q)(t|x)= \Phi(t, 0 | x) q(0 | x),
    \end{align}
    where $\Phi$ is the state-transition matrix, defined above in Equation \ref{eqn:statetransmat}. By definition 
   \begin{align}
       [\P q](t|x) &= \int_0^t \Phi(t, t_0 |x) q(t_0|x) \d t_0 \\
       &= U(t | x) \cdot \int_0^t U(t_0 | x)^{-1} q(t_0 | x) \d t_0 \\
       &= U(t|x) \cdot [V(U^{-1} q)](t|x) \\
       &= [(U \cdot V \cdot U^{-1}) q](t|x). 
   \end{align}
\end{proof}

In a discrete model, $U(t_i | x)$ is just a product of the Jacobian from the first time to time $t_i$. The Volterra integral operator on each trial just behaves by summing times up to the current time (equivalent to \texttt{torch.cumsum} over time axis): 
\begin{align}
    \tilde U(t_i | x) &= J_z(t_i | x) \cdot J_z(t_{i-1} | x) \cdots J_z(0 | x), \, \, \, \, \, \, \tilde V = \Delta t \cdot \begin{pmatrix}
        1 & 0 & 0 & \dots \\
        1 & 1 & 0 & \dots \\
        1 & 1 & 1 & \dots \\
        \vdots & \vdots & \vdots & \ddots
    \end{pmatrix}
\end{align}

\section{Experimental Details and Additions}

\subsection{Experiment 1}
\label{sec:apex1}

\begin{figure}[H]
    \centering
    \includegraphics[width=0.4\linewidth]{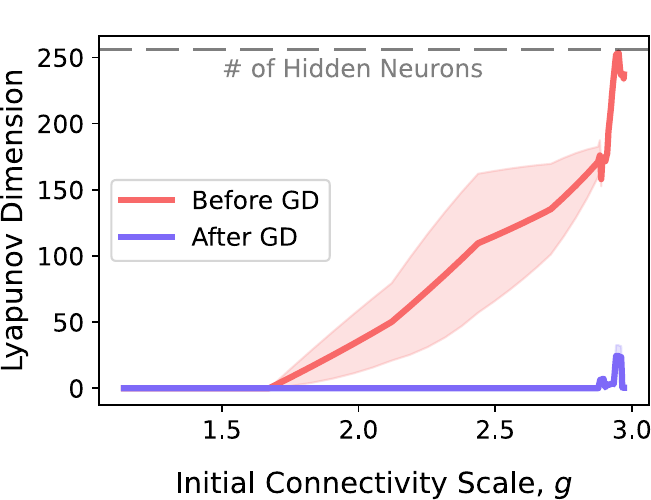}
    \caption{\textbf{Lyapunov dimension (Kaplan-Yorke dimension) \citep{frederickson1983liapunov}, corresponding to the sweep in Figure 2 of the main text, with networks evaluated on the Memory-Pro task.} Note that initial scale $g = 3$ leads to a maximal Lyapunov dimension of 256, which was the number of hidden units. Shaded regions correspond to variance between GD re-runs.}
    \label{fig:lyapdim}
\end{figure}

\paragraph{Detailed Setup} All models consisted of 256 hidden units. 500 models were trained and those with a loss below a threshold of $0.01665$ were said to have converged, found experimentally by taking $0.05$ times the final loss of a model with $g = 0$, as in prior work \citep{NEURIPS2020sch}. The data consisted of 3000 copies of trials from the memory pro task with white noise added with a mean of zero, as defined in \verb|custom_tasks/memory_pro.py| in our code. A batch size of 200 was used with the ADAM optimizer, which uses gradients internally that agree with our operator formulation (although the addition of acceleration, adjusting the learning rate, could make ADAM's learned solutions differ from the true gradient flow with an infinitesimally small learning rate, which is something to consider in more detail in future work). The learning rate was 0.001 and we applied gradient clipping with a threshold of 0.0001 to avoid massive gradients. When evaluating the model as in the paper, white noise was removed from the testing set of task trials. 

\paragraph{Lyapunov Dimension} In Supplementary Figure \ref{fig:lyapdim} we show the Lyapunov dimension corresponding to the sweep in the main text show in Figure 2B-C. This is calculated as the Kaplan-Yorke dimension \cite{frederickson1983liapunov}. Note that $g = 3$ leads to maximal Lyapunov dimension, hence why we chose this as the final $g$ value for the sweep.

\paragraph{Adjoint-Kernel Alignment is Maximized} We experimentally found that the adjoint signal, $\P^* \Err$, which is initially completely unaligned with the parameter operator, $\K$, rapidly became aligned with this operator. In particular, the Rayleigh quotient $R(\K, \P^* \Err)$ typically monotonically increased as the loss decreased in almost all cases considered (see description of this measure below in \ref{sec:apex2}). Understanding why this alignment maximization occurs is a future direction; here, we note that it is advantageous for faster learning. In particular, the adjoint signal $a = \P^* \Err$ is directly filtered through $\K$ to produce eventual changes to the hidden state $z$. If $a$ is not well aligned with $\K$, it can almost entirely be zeroed out. On the other hand, if $a$ projects well onto the dominant eigenfunctions of $\K$, we will get larger adjustments and potentially faster learning. 

\begin{figure}[H]
    \centering
    \includegraphics[width=\linewidth]{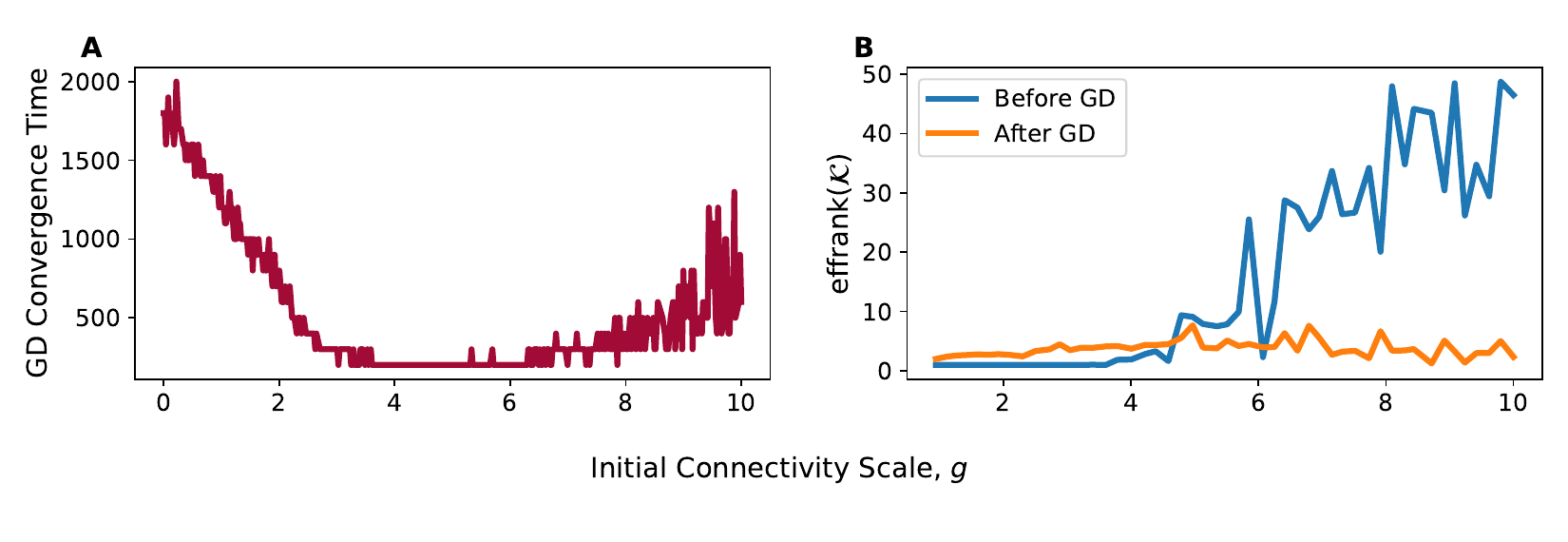}
    \caption{\textbf{Reproduction of the sweep in Experiment 1 (Section 3.1) of the main paper, varying initial connectivity scale, $g$, with GRU neural networks instead of RNNs for the Memory-Pro task.} \textbf{A} Convergence time in GD iterations, which almost monotonically decreased as the initial connectivity scale, $g$, increased. \textbf{B} Effective rank of $\K$ before and after training with varied $g$.}
    \label{fig:gruex1}
\end{figure}

\paragraph{GRU Case} We reproduced Experiment 1 from the main text (Section 3.1) using GRUs. In particular, since our theory applies to any general recurrent model, not just RNNs, we also reproduced Experiment 1 using 500 GRU neural networks with 256 units. In this case, we varied the initial weight scale $g$ of \verb|weight_hh_l| from $0$ to $10$ in the Pytorch implementation \citep{paszke2019pytorchimperativestylehighperformance}. We found that the wider range $0$ to $10$ was more applicable for the GRU, since it was more regular for smaller values of $g$ and exhibited a transition to more chaotic behavior around $g = 6$ (see Supplementary Figure \ref{fig:dimsgru}). The sweep results are shown in Supplementary Figure \ref{fig:gruex1}. In particular, the effective rank of $\K$ was much larger for large $g$ than for the RNN, however this rank was still much lower than the $\P$ operator. Note this rank is of the SVD averaged over hidden units, as described above in 2.3. We used $90$ timesteps per trial, and we evaluated on 30 trials, so the maximal rank of this operator is 2700. Thus, the $\K$ operator was still effectively a bottle-neck, but, unlike the RNN, the rank was much larger for high $g$, indicating more degrees of freedom during learning. Larger choices of $g$ led to faster learning, as in the RNN case, but only up to around $g =6$, after which increasing $g$ led to slower learning.

\begin{figure}[H]
    \centering
    \includegraphics[width=0.8\linewidth]{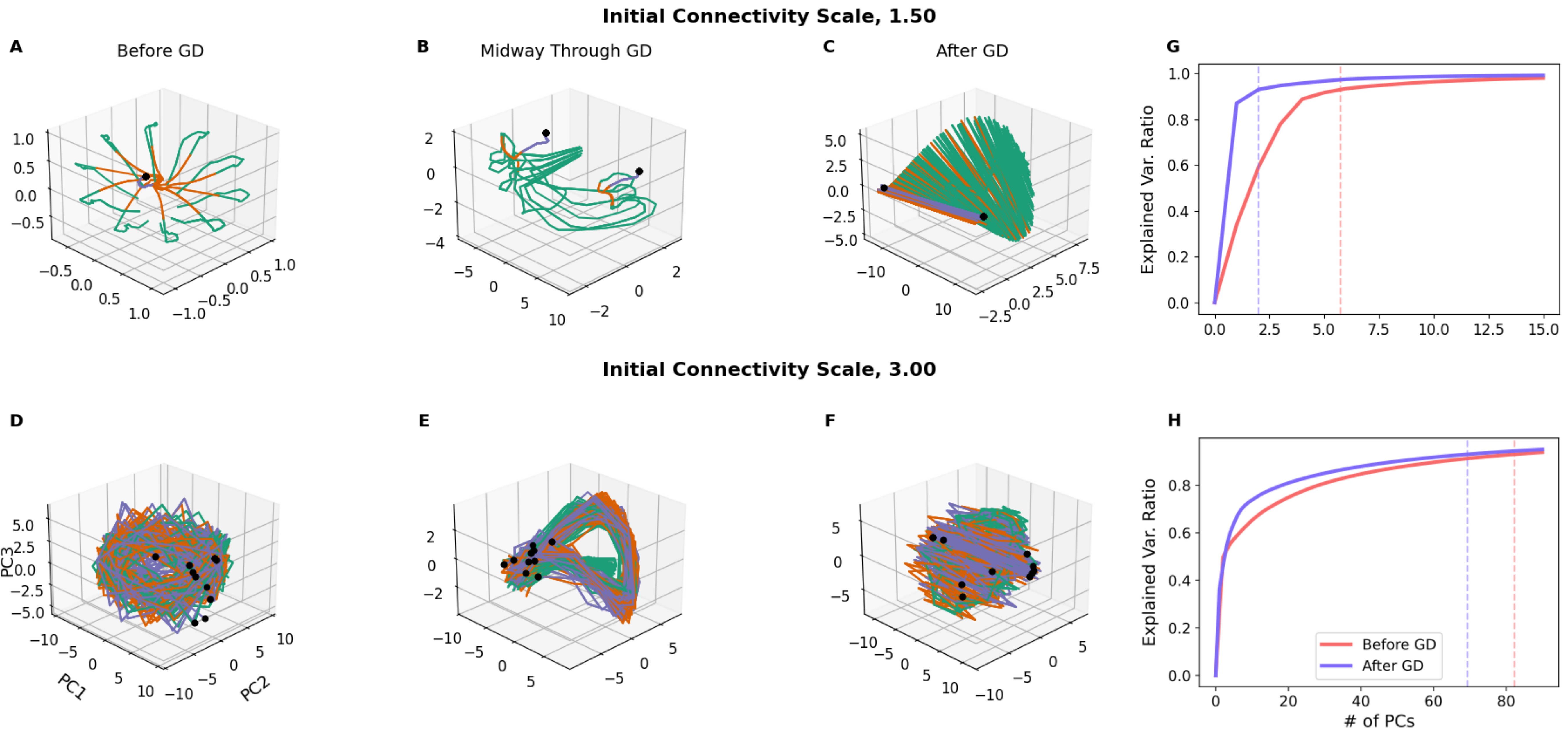}
    \caption{\textbf{Example learned dynamics for the Memory-Pro task, replacing tanh activation in main paper (Experiment 1, section 3.1) with ReLU.} Models were trained with GD until they reached a low final loss, below the threshold described in the Detailed Setup above. \textbf{A-C, D-F} Example hidden state trajectories over three GD iterations for $g = 1.5$ and $g = 3.0$ (top and bottom rows, respectively). Axes are defined by principal components (PCs), computed distinctly for each plot. Each panel shows the hidden state over different task inputs, colored by task period (green = Stimulus, orange = Memory, purple = Response). Black dots denote the final state $z(t_{end})$. \textbf{G, H} Dimensionality of hidden state dynamics before and after training for the two examples, with effective rank (number of PCs required to capture 95 percent variance) indicated.}
    \label{fig:relu}
\end{figure}

\paragraph{ReLU RNNs} Finally, we experimented with using ReLU activations instead of tanh with the RNN. The tanh activation implicitly regularizes dynamics since once the hidden state is sufficiently large in norm it behaves almost constant, with very small slow. It is also bounded above and below by $1$. On the other hand, ReLU is not bounded and behaves like a linear RNN when confined the quadrant where all components of the hidden state are positive. We assumed that this would lead to more complex or chaotic dynamics, in turn leading to a higher dimensional $\K$ operator. The dynamics of the final learned solution were indeed often more chaotic, as is consistent with prior work \citep{Tolmachev2024.11.23.625012}, and we found that the effective rank of $\K$ was also larger for the more chaotic, high $g$ and high-dimensional, learned solutions. This suggests that learning with ReLU activation exhibits less bottle-necking due to the rank of $\K$, leading to the observed chaotic, high dimensional dynamics. However, the dimension of $\K$ was still lower than that of $\P$ in all cases, which was typically close to the maximal effective rank. In the future, concocting models that maximize the rank of $\K$, making $\K$ as close as possible to the identity operator, which might prevent collapse to dynamic low-dimensional motifs, could be an insightful direction.

\subsection{Experiment 2}

\label{sec:apex2}

\paragraph{Detailed Setup} The same training hyperparameters were used as in Experiment 1 above, except that the memory-pro task was replaced by the four tasks memory-pro, memory-anti, delay-pro and delay-anti. 20 GRUs were trained for 5000 GD iterations and all converged well before the final iteration. 

\paragraph{Rayleigh Quotient Alignment}
Experiment 2 (Section 3.2) in the main text presents an ``interference matrix'' measuring how individual sub-tasks in a multi-task context align together. Here, we offer another matrix related to this interference matrix but based on Rayleigh-quotient operator alignments. In particular, an input per-trial trajectory produces large output from the operators considered when it is well aligned with the top SVD eigenfunctions of these operators. We can quantify this alignment by projecting a per-trial trajectory, e.g. the adjoint or hidden state, onto the operator's eigenfunctions by seeing how many of these we need to account for some percentage of the variance in the original signal. Alternatively, a more efficient way to measure alignment is with the Rayleigh quotient. For example, for the operator $\K$ and an input per-trial trajectory, $q \in \T$, this is computed as
\begin{align}
    R(\K, q) := \frac{\langle \K q, q \rangle}{\langle q, q \rangle}.
\end{align}
The inner product takes a trace over corresponding times, trials and hidden units. In particular, $\langle q, q \rangle$, for example, can be computed by taking the component-wise product of $q$ with $q$ and then taking a mean over all trials and hidden units and summing (or integrating in a continuous context) over all times. The quantity $R(\K, q)$ measures how much $\K$ aligns with the signal $q$. To define a metric based on this, as in Experiment 2 in the main text, we can measure how the operator $\K_{ij}$, measuring interference from sub-task $j$ on the gradient flow for the restricted sub-task $i$ hidden states, aligns with the sub-task $j$ adjoint $\P_j^* \Err_j$, inputted into this operator. In particular, we can form a scalar matrix $M_{ij} = R(K_{ij}, \P^*_j \Err_j)$. Similar metrics can be defined using the Rayleigh quotient. Our code implements the Rayleigh quotient, and future work could possibly use it to extend our analysis tools. 

\begin{figure}[t]
    \centering
    \includegraphics[width=.6\linewidth]{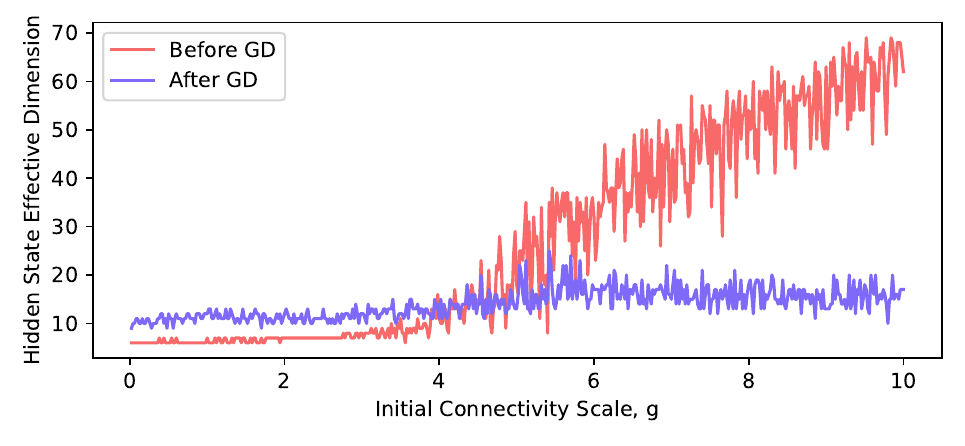}
    \caption{\textbf{Effective dimension of GRU generated hidden activity on multi-task problem in Experiment 2 before and after training.} As in the RNN, this dimension grows with initial weight scale, $g$, but we find that a large range (0-10 instead of 0-3) is needed with the GRU to produce similar trends as in the RNN case, with maximal Lyapunov dimension only reached for $g$ around 10.}
    \label{fig:dimsgru}
\end{figure}

\begin{figure}[t]
    \centering
    \includegraphics[width=.9\linewidth]{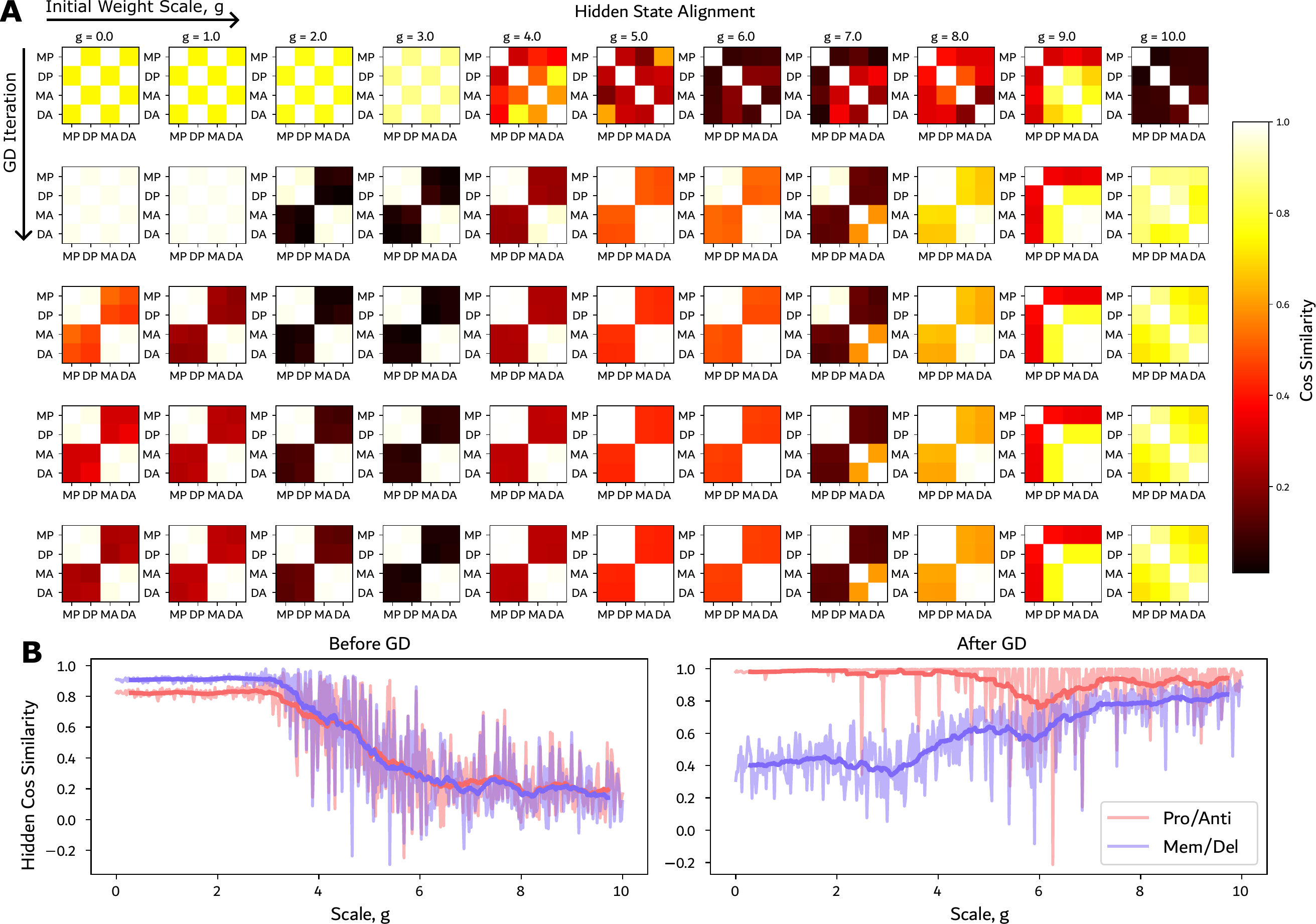}
    \caption{\textbf{Emergence of shared dynamics for the four-task context of Experiment 2 in the main text.} \textbf{A} Cosine similarity between the hidden state representations with identical stimulus inputs for each. Before training, for low initial recurrent weight scales, $g<3$, similar input timecourses result in preferential alignment into the Pro/Anti organization, in which the Memory-Pro and Delay-Pro tasks were highly aligned, and likewise the Memory-Anti and Delay-Anti tasks were aligned. For large $g$, more diverse dynamics emerged without as clear of a bias towards the Pro/Anti sharing. \textbf{B} Comparing hidden state alignment into the two Mem/Del and Pro/Anti organizations before and after training, confirming the observations in \textbf{A}. Note bold lines are given by filtering with a window over 10 consecutive runs and faded lines show the true, noisier values. }
    \label{fig:gsweepex2}
\end{figure}

\subsection{Varying Initial Weight Scale in Experiment 2}

\label{sec:apex3}

It is natural to extend the methods of Experiment 1 to Experiment 2 by considering different initial scale values, $g$, in the multi-task context considered. To do so, we varied the recurrent weight scale, $g$, for the GRU for 500 models with $g$ between $0$ and $10$. The results are summarized in Supplementary Figure \ref{fig:gsweepex2}. As in the GRU extension of Experiment 1 above in this Appendix, we found that a wider range of $g$ values is appropriate with the GRU compared to the RNN, likely due to regularity of this model, with high-dimensional chaotic behavior emerging around $g = 6$, instead of around $1$ to $1.5$ with the RNN (see dimensions in Supplementary Figure \ref{fig:dimsgru}). We found that learning for all GRUs converged with $g$ in this wide range. Sub-task alignment based on cosine-similarity was similar to the ones in Experiment 2 of the main text for $g$ in the range around $0$ to $6$. For $g > 6$, task organization was less predictable and more tasks appeared to have shared subspace structure. 

As in Experiment 2 in the main text, we measured how tasks aligned, specifically considering how tasks were aligned either based on type, Mem/Del, meaning that Memory-Pro and Memory-Anti form well-aligned attractors, and so too do Delay-Pro and Delay-Anti, or, alternatively, Pro/Anti, where Memory-Pro and Delay-Pro are well aligned, and so too are Memory-Anti and Delay-Anti (see Figure 4E). Alignment was measured through the hidden state cosine similarity of all trajectories with identical input stimuli. We found that a consistent bias towards strongly coupled Pro/Anti organization appeared for smaller $g$ values, as in the main text. With large $g$ the patterns of attractor sharing were less predictable (see Supplementary Figure \ref{fig:gsweepex2}). In this case, we found that the alignment between tasks was more distributed, without a clear bias towards the Pro/Anti organization, even though the final learned hidden state dimension was similar in effective dimension (see Supplementary Figure \ref{fig:dimsgru}). To explain this trend in the context of the KPFlow operators, we measured the cumulative cosine similarity of the $\delta z$ objectives, using the interference matrix introduced in the main text, explicitly using the operators associated with each sub-task. Examples of this cumulative alignment for all $g$ values and at different GD iterations are shown in Supplementary Figure \ref{fig:gsweepex2part2}. In particular, after 100 iterations the Pro/Anti organization showed substantially higher alignment for smaller $g$. For larger $g$, the overall $\delta z $ objective alignment was smaller and showed less of a distinction between Mem/Del and Pro/Anti, especially at earlier iterations. 
Thus, we find that during the initial training of the GRU model for large $g$ there are chaotic, less aligned updates existing in a high-dimensional activity space, leading to less of a bias towards Pro/Anti. As the task is learned, the latent activity dimension reduces and the updates become more aligned, but overall there is a higher diversity of configurations learned with large $g$. We suspect that small $g$ results in reuse of attractors across similar tasks due to alignment of $\delta z$; however, task alignment for large $g$ must be more related to the loss landscape in a chaotic dynamical system. These two regimes should be investigated further.

%
\begin{figure}[t]
    \centering
    \includegraphics[width=\linewidth]{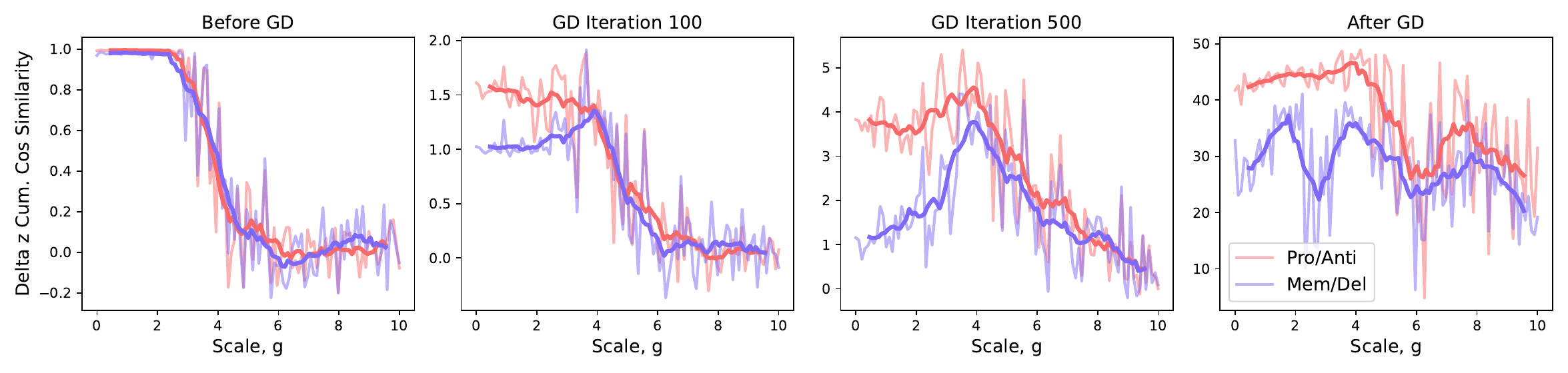}
    \caption{\textbf{Cumulative alignment of objectives related to four sub-tasks, as in main text, filtered according to Mem/Del or Pro/Anti organizations.} Each snapshot shows the cumulative alignment of objectives at a particular GD iteration. Note a few iterations into GD, the Pro/Anti organization is more present in the delta z cumulative updates compared to the Mem/Del for small $g$. For large $g$, the overall delta z alignment is lower and there is less of a distinction between the two organizations, leading to the representations shown in Supplementary Figure \ref{fig:gsweepex2}. Bold lines show values filtered with a sliding window of 10 consecutive runs, while transparent lines show the raw values. } 
    \label{fig:gsweepex2part2}
\end{figure}


\subsection{Hodgkin-Huxley Network}
\label{sec:hhappend}
In this section, to show a diverse example application of our decomposition, we work through the derivation of the operators involved in our formulation for a network of biophysical Hodgkin-Huxley (H-H) neurons \citep{hodgkin1952quantitative}. This acts as a first step towards building a rigorous understanding of how learning takes place under GD in such networks. Prior work has shown that GD on such networks can exhibit many pitfalls \citep{hazelden2023evolutionary}. 

 Suppose our state space is $S = \R \oplus [0, 1]^3 \subset \R^{4n}$ with 
\begin{align}
    z(t) = \begin{bmatrix}
        V(t) \\ m(t) \\ n(t) \\ h(t)
    \end{bmatrix}, \, \, V(t) \in \R, \, m, n, h \in [0,1]. 
\end{align}
The dynamics are given by task trial input $x(t)$ applied to the neurons. The trained parameters are assumed to be two weights $W, W_{in}$ and the H-H parameters are set to be constant for simplicity:
\begin{align}
    \frac{\d}{\d t} z(t) = f(z(t)  | x; \{W_{in}, W\}) = \begin{bmatrix}
        f_V(V(t) | x; I_{app}, W_{in}, W) \\
        f_m(m(t), V(t)) \\
        f_n(n(t), V(t)) \\
        f_h(h(t), V(t))
    \end{bmatrix} \in \R^{4n},
\end{align}
with
\begin{align}
    \frac{\d V(t|x)}{\d t} &= g_K n^4 \odot (V-V_k) + g_{Na} m^3 \odot h \odot (V - V_{Na}) + g_l (V - V_l) + W \sigma(V) + W_{in} x(t) + I_{app}, \\
    \label{eqn:gating}
    \frac{\d k}{\d t} &= \alpha_k(V) \odot (1- k) - \beta_k (V) \odot k, \,  \, \, \, \text{ for } k = m, n, h.
\end{align}
The function $\sigma$ is a component-wise function that measures ``output'' of the network and can be discrete (e.g. a Heaviside) in which case we need to use the distributional derivative or ``surrogate gradients'' \citep{neftci2019surrogategradientlearningspiking}. For example, we can choose $\sigma = \text{sigmoid}((V - V_t) / K_p)$ for some constant $V_t, K_p$ \citep{hazelden2023evolutionary}. The functions $\sigma_k, \beta_k$ are also component-wise with well defined sigmoid-like form depending on the variant of Hodgkin-Huxley neuron we want to model \citep{gerstner2014neuronal}. The parameters $g_k, g_{Na}, g_l, V_k, V_{Na}, V_l$ can also be specified per-neuron in which case the multiplication is component-wise. 

Note in the above that the $|x$ notation (dependence on the trial input $x$) is suppressed for succinctness. In reality, we simulate the network with many distinct inputs $x \sim X$, as in the main text. We can write the dynamics in the alternative form 
\begin{align}
    \frac{\d V(t|x)}{\d t} &= G \odot V - E + W \sigma(V) + W_{in} x(t) + I_{app}, \\
    \text{ where } \, G(z) &= g_{K} n^4 + g_{Na} m^3 \odot h + g_l, \\
    E(z) &= g_k V_k n^4 + g_{Na} V_{Na}  m^3 \odot h + g_l V_l. 
\end{align}

\paragraph{State-Transition Matrix} 
The operator $\P$ is defined by the state-transition matrix $\Phi(t, t_0 | x)$, as in \ref{eqn:statetransitiondef}. From the above, we see that the Jacobian has structure:
\begin{align}
    D_z f &= \begin{bmatrix}
        D_V f_V & D_m f_V & D_n f_V & D_h f_V \\
        D_V f_m & D_m f_m & 0 & 0 \\
        D_V f_n & 0 & D_V f_n & 0 \\
        D_V f_h & 0 & 0 & D_V f_h
    \end{bmatrix} \in \R^{4n \times 4n},
\end{align}
where
\begin{align}
    D_V f_V &= \diag(g_k n^4 + g_{Na} m^3 \odot h + g_l)  + W \diag(\sigma'(V)), \\
    D_V f_k &= -\diag((\alpha_k'(V) + \beta_k'(V)) \odot k) \\
    D_k f_k &= -\diag(\alpha_k(V) + \beta_k(V)), \,  \, \, \, \, \, \, \, \, \, \, \, \, \, \, \, \, \text{ for } k = m, n, h. 
\end{align}
Finally,
\begin{align}
    D_m f_V &= \diag(3 g_{Na} m^2 \odot h \odot (V - V_{Na})), \\
    D_n f_V &= \diag(4 g_{K} n^3 \odot (V - V_{K})), \\
    D_h f_V &= \diag(g_{Na} m^3 \odot (V - V_{Na})).
\end{align}
Note that the matrix $D_z f$ is a ``block arrowhead matrix'' where the only non-diagonal block is $D_V f_V$, which is not diagonal due to the weights $W$. Unfortunately, multiplying such matrices together can result in a fully-dense matrix. In particular, the state transition matrix can be written in the form 
\begin{align}
    \Phi(t, t_0 | x) = \begin{pmatrix} \Phi_{V,V} & \Phi_{V, m} & \Phi_{V, n} & \Phi_{V, h} \\
                                       \Phi_{m, V} & \Phi_{m,m} & \Phi_{m, n} & \Phi_{m, h} \\
                                       \Phi_{n, V} & \Phi_{n,m} & \Phi_{n, n} & \Phi_{n, h} \\
                                       \Phi_{h, V} & \Phi_{h,m} & \Phi_{h, n} & \Phi_{h, h}
                        \end{pmatrix}(t, t_0 | x),
\end{align}
where each term is an $n$ by $n$ block matrix. Then, using the Jacobian above, the dynamics of each block are 
\begin{align}
    \frac{\d}{\d t} \Phi_{V,j}(t, t_0 |x)&= (\diag(g_k n^4 + g_{Na} m^3 \odot h + g_l)  + W \diag(\sigma'(V))) \Phi_{V,j}(t,t_0|x) \\
                                     &+ \diag(3 g_{Na} m^2 \odot h \odot (V - V_{Na})) \Phi_{m, j}(t,t_0 | x)  \\
                                     &+ \diag(4 g_{K} n^3 \odot (V - V_{K})) \Phi_{n, j}(t, t_0 | x) \\
                                     &+ \diag(g_{Na} m^3 \odot (V - V_{Na})) \Phi_{h, j}(t, t_0 | x), \, \, \, \text{ for } j = V, m, n, h,
\end{align}
and
\begin{align}
    \frac{\d}{\d t} \Phi_{k, j} (t, t_0 | x) &= -\diag((\alpha_k'(V) + \beta_k'(V)) \odot k) \Phi_{V,j}(t, t_0 | x) \\
    &- \diag(\alpha_k(V) + \beta_k(V)) \Phi_{k, j} (t, t_0 | x), \, \text{ for } k = m,n,h; j = V, m , n, h,
\end{align}
which represents a decay to zero (since $\alpha_k(V) + \beta_k(V) > 0$ always) driven by the non-diagonal term $\Phi_V(t, t_0 | x)$. So, the voltage related terms $\Phi_{V, k}$ in the first row have consistent dynamics, while the terms in the rows below, $\Phi_{k, j}$ also share their own consistent dynamics representing a decay to zero driven by the corresponding row one voltage $\Phi$ term. Initially, since the state-transition matrix starts as the identity, we have $\Phi_{j,j}(t_0, t_0 | x) = \text{Id}_{n \times n}$ for $j= V, m, n, h$ and all other terms are zero.

Thus, in summary, we see that the linearized flow determined by the state transition (see \ref{eqn:statetransitiondef}) has similar characteristics as the original H-H dynamics, with the state-transition matrix being broken into dynamics consistent for the first, voltage related, row, and dynamics consistent with the other rows related to changes to the gating variables. 

\paragraph{Parameter Kernel} \text{ }

The parameter kernel is computed through the Jacobian term $J_\theta := D_\theta f $. First, we compute $D_{W_{in}} f$, given by,
\begin{align}
    D_{W_{in}} f_k &= 0 , \, \, \, k = m,n,h, \\
    D_{W_{in}} f_V &= x(t)^T. 
\end{align}
Rigorously, $D_{W_{in}} f_V = x(t) \otimes \text{Id}_{n \times n_{in}}$ is a three tensor, however, the action of this tensor agrees with $x(t)^T$, so we use this notation for simplicity. So, effectively,
\begin{align}
    D_{W_{in}} f &= \begin{bmatrix} x(t)^T & 0 & 0 & 0\end{bmatrix},
\end{align}
which is a row vector in $\R^{4n}$. Likewise, 
\begin{align}
    D_{W} f &= \begin{bmatrix} \sigma(V(t|x))^T & 0 & 0 & 0 \end{bmatrix}.
\end{align}
Thus, the kernel operator defining the inner product induced parameter kernel $\K$ in our work is given by
\begin{align}
    k(t_0, t_1 | x_0, x_1) := D_\theta f(t_0 | x_0) D_\theta f(t_1 | x_1)^T &= x_0(t_0)^Tx_1(t_1) +  \sigma(V(t_0 | x_0))^T\sigma(V(t_1 | x_1)).
\end{align}
Note that if $\sigma$ quantifies ``neuron is spiking,'' e.g. $\sigma(V) = 1$ if it is spiking and $\sigma(V) = 0$ if not, then the inner product is equal to the number of neurons, $i$, spiking at time $t_0$, $t_1$ on \textit{both} trials $x_0, x_1$. In symbols, if $\delta^{sp}_i(t|x)$ means ``$i$ is spiking at time $t$ on trial $x$,''
\begin{align}
    \sigma(v_i) = \delta^{sp}_i \, \Rightarrow \langle \sigma(V(t_0 | x_0)), \sigma(V(t_1 | x_1))\rangle = \# \{i | \delta_{i}^{sp}(t_0|x_0) = \delta_{i}^{sp}(t_1 | x_1) = 1\}.
\end{align}
Thus, we can write (similar to the RNN example in the main text and Theorem 1), $\K = \K_1 + \K_2$ where $\K_1, \K_2$ are kernel operators induced by learning the weights $W_{in}, W$, respectively. In particular, the operator $\K_2$ measures correlation between spikes trains over all times, hidden neurons and input trials. Thus, this kernel, weighting by how well aligned the spike trains are, is reminiscent of ideas from Hebbian learning in which neurons that are more correlated together have their synapses promoted. 

\paragraph{Summary} We have derived the state-transition dynamics and the parameter operator for the network of H-H neurons described. We provide this as a first step illustration of the applicability of our approach to more diverse models. Future work could build on this derivation, for example using approximations such as the Magnus expansion of the state-transition matrix to simplify the parameter operator \citep{nijmeijer1990nonlinear}. Furthermore, implementations of the H-H network described can be directly analyzed throughout GD training using the operator-based analysis tools we provide in our \texttt{KPFlow} software. Certain terms, e.g. $\alpha_k(V) + \beta_k(V)$ above, which is always positive, could give rise to a better understanding of how the state-transition dynamics evolve during feed-forward evaluation of the H-H network.